\documentclass[aps,pre,twocolumn,showpacs,superscriptaddress]{revtex4}
\usepackage{amsmath}
\usepackage{amssymb}
\usepackage{epsfig}
\usepackage{color}
\usepackage{graphicx}
\usepackage{graphicx,amsmath}
\usepackage{bm}
\usepackage{hyperref}

\begin{document}
\newcommand{\beq}{\begin{equation}}
\newcommand{\eeq}{\end{equation}}
\title{Breast Cancer Diagnosis by Higher-Order Probabilistic Perceptrons}
\author{Aditya Cowsik}
\affiliation{Physics Department, Princeton University, Princeton, NJ 08540} 
\author{John W.~Clark}
\affiliation{McDonnell Center for the Space Sciences \& Department
Physics, Washington University, St.~Louis, MO 63130, USA}
\affiliation{
Centro de Investiga\c{c}\~{a}o em Matem\'{a}tica e Aplica\c{c}\~{o}es,
University of Madeira, 9020-105 Funchal, Madeira, Portugal
9000-390 Funchal, Madeira, Portugal} 
\begin{abstract}
	
A two-layer neural network model that systematically includes correlations 
among input variables to arbitrary order and is designed to implement Bayes 
inference has been adapted to classify breast cancer tumors as malignant 
or benign, assigning a probability for either outcome. The inputs to the 
network represent measured characteristics of cell nuclei imaged in Fine 
Needle Aspiration biopsies. The present machine-learning approach to 
diagnosis (known as HOPP, for higher-order probabilistic perceptron) is 
tested on the much-studied, open-access Breast Cancer Wisconsin 
(Diagnosis) Data Set of Wolberg et al.  This set lists, for each tumor, 
measured physical parameters of the cell nuclei of each sample.  
The HOPP model can identify the key factors -- input features and 
their combinations -- most relevant for reliable diagnosis.
HOPP networks were trained on 90\% of the examples
in the Wisconsin database, and tested on the remaining 10\%.
Referred to ensembles of 300 networks, selected randomly for cross-validation, 
accuracy of classification for the test sets of up to 97\% was readily 
achieved, with standard deviation around 2\%, together with average 
Matthews correlation coefficients reaching 0.94 indicating excellent predictive 
performance. Demonstrably, the HOPP is capable of matching the predictive power 
attained by other advanced machine-learning algorithms applied to this 
much-studied database, over several decades. Analysis shows that in this 
special problem, which is almost linearly separable, the effects of irreducible 
correlations among the measured features of the Wisconsin database are of 
relatively minor importance, as the Naive Bayes approximation can itself 
yield predictive accuracy approaching 95\%. The advantages of the HOPP 
algorithm will be more clearly revealed in application to more
challenging machine-learning problems.

\end{abstract}


\maketitle

\section{Introduction}

We shall describe a new machine-learning technique for breast-cancer 
diagnosis having a basis in Bayesian probability theory that not
only achieves accurate classification of fine-needle aspiration 
biopsies but also (i) directly yields a prediction of the probability 
of malignancy and (ii) provides for systematic analysis of the effects 
on this prediction of irreducible correlations among the measured 
features of cell nuclei.  The motivation for such research stems from 
the sheer magnitude of the problem posed by breast cancer: Among the 
various health and medical problems confronting our society to-day 
breast cancer is one of the most serious.  The statistical data 
\cite{bcstatis} released by the American Cancer Society show that 
except for skin cancer, breast cancer has the highest frequency of 
occurrence among women, apart from skin cancer.  About 12\% of 
women in the US alone will develop invasive breast cancer at some time 
in their lives, and during the year 2019 it is estimated that 268,000
women will have received this diagnosis. There are more than 3 million 
breast cancer survivors in the US who have to be monitored regularly 
for recurrence \cite{bcstatis}.  

Even though regular screening through physical examination and mammograms 
of most women above the age of 35 $\sim$ 40 is being conducted, a 
definitive diagnosis of any discovered lesion or tumor requires cytological 
study of the cells extracted from the tumor. Fine Needle Aspiration 
Biopsy (FNAB) is the least invasive procedure in support of such a 
diagnosis \cite{yu}: A very fine needle is inserted into the cyst or tumor 
to remove a small sample of fluid, cells, and tissue, which is then 
transferred to a slide and prepared for examination under a microscope. 
A set of such slides is examined carefully by an expert pathologist, 
noting various features such as agglomerations of cells and distortions 
in the shapes and sizes of cell nuclei as a basis for diagnosis. Digital 
photography and computer-aided image processing facilitate this process. 
The magnitude of the challenge faced by the medical profession in 
diagnostics and treatment is scaled not only by the prevalence of the 
disease and the difficulties of treatment, but also by the enormous 
number of biopsies that must be performed and evaluated.  

One of the earliest successes in meeting the latter challenge with emerging 
classification technologies is that of Wolberg et al.~\cite{wolberg90,wolberg93a,wolberg93b,wolberg94,wolberg95a,wolberg95b,dataset}, 
who applied digital image analysis together with a variant of the 
MSM-Tree multi-surface method \cite{mang68,mang93,bennett92a,bennett92b} 
to achieve automatic classification of tumors as benign or as malignant on 
the basis of FNAB biopsies.  This pioneering step has been followed by 
a prodigious flow of research in the same spirit, involving the development 
and application of artificial neural networks and other automated 
learning systems to breast-cancer diagnosis 
\cite{dawson,maclin1,maclin2,golberg,wilding1,wilding2,ravdin1,ravdin2,ravdin3,nafe,wu,rogers,floyd,roy,fogel1,fogel2,sahiner,setiono,burke,furundzic,pena,setiono2,west,abbass,aragones,meesad,chen,kiyan,chou,nahar,elizondo,uberli,tcchen,jelen,huang,akay,subashini,karabatak,liang,paulin09,paulin10,paulin11,peres,moradi,hlchen,cedeno,elgader,kowal,blachnik,george,aloraini,gu,savitha,alkim,ahchen,nahar2,utomo,zheng,dheeba,seera,bhardwaj,mert,nahato,karabatak2,kim,abdel,mohammed,chaurasia,chao}.  
Computer-assisted diagnosis of breast cancer has been reviewed
in Ref.~\cite{rangayyan}.  Computational studies comparing the 
performance of diverse classifier algorithms applied this problem 
may be found in Refs.~\cite{sadja,cruz,you,beg,asri,bazazeh,mandal}, 
as well as many of the targeted studies already cited.  For incisive 
commentary on the role of machine learning in medical diagnosis, see 
Ref.~\cite{kononrev}; for warnings about misuses of neural
networks classifiers for cancer diagnosis, see Ref.~\cite{schwarzer}.  
A survey of expert systems in their broader applications has been 
provided in Ref.~\cite{liao}, and more specifically to medical 
diagnosis, in Ref.~\cite{lisboa}.

The work reported here was undertaken in this context. Numerous computational 
strategies implemented previously in the research cited above have achieved 
a high level of accuracy in the diagnosis of breast cancer based on 
Fine-Needle Aspiration Biopsies (FNAB). While these methods are able to 
efficiently classify such biopsies as benign or malignant, there has been 
less attention has been given in determining, among the 10 to 30 
characteristics or features of cell nuclei measured in creating the 
Wisconsin database, those specific features -- {\it especially in
combination} -- that are most crucial for correct diagnosis.  Knowledge
of such incisive determinants could potentially yield new insights
into the underlying biophysical and biochemical factors responsible for
malignancy.  As will be seen, the extent to which this is practical
depends largely on the intrinsic difficulty of the classification
problem in question.

\section{Higher-Order Probabilistic Perceptron}

Over the past four decades many pattern-classification techniques and
algorithms have been developed by applied mathematicians, engineers,
and physicists \cite{duda,bishop,neal,vapnik,haykin,elomaa}.  Those based on 
``artificial intelligence'' as represented by feedforward neural network 
systems \cite{bishop}, Bayesian network models \cite{neal} , and 
support vector machines \cite{vapnik} are the most relevant to the 
present work. Having a set of input patterns whose classifications are known 
{\it a priori}, such systems can be trained to classify new input 
patterns having previously unknown class assignments. We intend not 
only to classify biopsies into two sub-classes: (a) benign and (b) 
malignant, but also to predict the probability that any sample is 
malignant. Morover, we will explore the prospects for identifying
the specific factors derived from FNAB that are most instrumental 
to such classification.

\begin{figure}
\includegraphics[width=.40\textwidth]{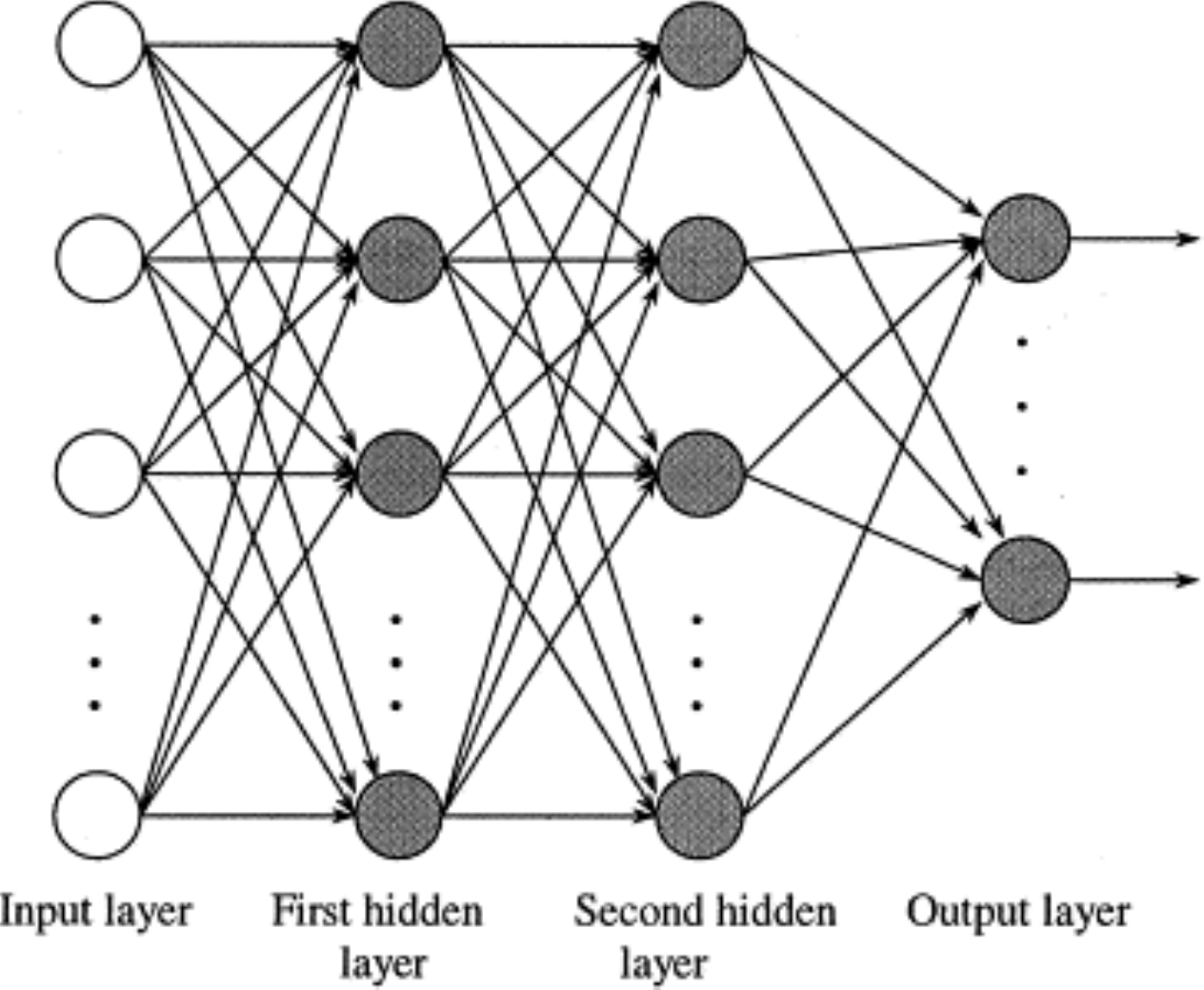}
\caption{Conventional multilayer feedforward neural network having
two intermediate layers of ``hidden neurons'' between input and output 
layers.  Darkened circles represent processing units analogous to 
neurons; lines oriented forward symbolize weighted connections between 
units analogous to interneuron synapses.  Information flows left to right 
as units in each layer simulate those in the next layer. Such networks
are commonly taught to recognize and classify input patterns via the 
backpropagation algorithm \cite{haykin}.}
\label{Contrast-figure}
\end{figure}

\begin{figure}
\vskip -2.3truein
\hspace*{-2cm}\includegraphics[width=.65\textwidth]{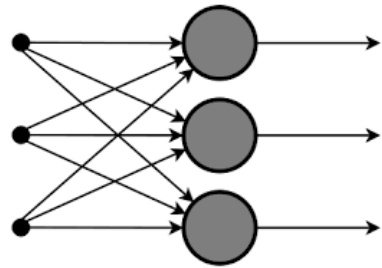}
\vskip -2.3truein
\caption{
Two-layer feed-forward architecture of the Rosenblatt's Elementary 
Perceptron \cite{rosenblatt} and the Higher-order Probabilistic 
Perceptron (HOPP) \cite{clark}.  Meaning of circles and lines is 
the same as in Fig.~1.}
\end{figure}

The unique properties of the higher-order probabilistic HOPP model 
developed by Clark et al.~\cite{clark} are especially suited to the 
task of identifying such crucial factors in a rather general category of 
classification problems.  Figs.~1 and 2 compare two familiar architectures 
for feedforward neural networks. In contrast to the rather opaque 
multilayer neural networks in extensive use, the HOPP model has the 
transparent architecture of the elementary two-layer perceptron 
\cite{rosenblatt}, without any hidden intermediate layers 
\cite{clark,clarkbook}.  The compensating 
complication relative to conventional neural-network classifiers is 
that the HOPP model must allow for the inclusion of all higher-order 
couplings of all subsets of input units to each output unit, not just the 
simple neuron-neuron couplings involving a single input neuron and a given
output neuron (along with single-neuron biases/thresholds), which
govern the dynamics of standard neural-network models.  Clark et al.\ 
have demonstrated that the HOPP architecture is sufficiently general 
for the full array statistical correlations among measured variables 
that is inherent to Bayesian inference for such a problem are incorporated.  
This means that the effects of correlations between the input variables 
on output decisions can, in practice, be made explicit through 
systematic determination of the higher-order couplings by training 
the HOPP network on a sufficiently representative database. 

The Elementary Perceptron is the ``primitive'' of feedforward neural-network 
architectures.  Generalization to the Multilayer Perceptron structure 
adds a dimension of {\it depth}, allowing for the representation of 
the data as a hierararchy of feature maps that are implemented
in successive layers.  The more layers, the deeper the network,
hence the current focus on {\it Deep Neural Networks}.  In the same
spirit, one may add ``width'' to the Elementary Perceptron by
enriching its single processing layer with higher-order couplings
of its neuron-like units, corresponding to the introduction, in the 
physics of many-body sytems, of three-body, four-body, $\cdots$, 
$K$-body interactions along with familiar two-body interactions  
provided, for example, by the Coulomb potential.  

Naturally, one may also add {\it width} in different way to Multilayer
Perceptrons by increasing the numbers of hidden neurons in intermediate 
layers. Moreover, an even more general architecture may be created
by allowing for higher-order couplings of the model neurons in Deep Neural 
Networks.

\section{HOPP Model with Continuous Inputs}

In order to classify of the breast tumors of the Wisconsin Breast Cancer 
Database (WBCD), we have trivially adapted the HOPP model to accept 
continuous inputs, in such a way as to exclude self-correlations of 
arbitrary order.  The absence of hidden intermediate layers between 
the input and the output layers allows us to explicitly access both 
the direct effects of specific inputs and the effects of all correlations 
among them, in principle to all relevant orders, after completion 
of the training process. One may well be concerned that with inputs 
corresponding to 10 to 30 characteristics, the total number of coupling 
coefficients (``weights'') in the network will become too large with 
increasing correlation order for this approach to be useful, because 
this tends to cause overfitting at the expense of generalization.  
However, the HOPP model admits effective pruning procedures \cite{clark}) 
that can operate during the training process to limit the final set 
of couplings to a few 10's in the present case, while still achieving 
reliable classification of new cases.  In the general context of medical 
diagnosis, success in identifying a reduced set of weights as the 
most salient salient might provide valuable clues to the nature of the 
underlying mechanisms, potentially at the molecular level, of the illness 
or malfunction in question.

A continuous-input version of the HOPP model suitable for classification 
of breast tumors features an input layer of $K$ units having real-value 
activities $x_1,x_2,\ldots,x_K$, one such unit for each of the measured 
features of the cell nuclei of the biopsy.  Each of $L=2$ output 
units $\lambda$, which respectively signal malignant and benign conditions, 
receives stimuli from all possible {\it combinations} of input units, 
in the form of a product of the activities of the inputs involved, weighted 
by a real-number factor specific to each combination.  Explicitly, the 
un-normalized output of processing unit $\lambda$ is given by
\begin{eqnarray}
u_{\lambda}(\textbf{x}) &=& w_{\lambda,0} + \sum_{i=1}^{K} w_{\lambda, i}x_{i} 
+ \sum_{i < j} w_{\lambda, ij}x_{i}x_{j} \nonumber \\ 
&+& \sum_{i < j < k} w_{\lambda, ijk}x_{i}x_{j}x_{k} 
+ \cdots + \nonumber \\ 
&+& \sum_{i_1 < i_2 \ldots < i_K} 
w_{\lambda, i_1 i_2 \ldots i_K} x_{i_1}x_{i_2}\ldots x_{i_K}, 
\label{uexp}
\end{eqnarray}
which includes a bias parameter $w_{0,\lambda}$ assigned to output 
$\lambda$.

This expression can be evaluated for any input vector $\textbf{x}
=(x_1,x_2,\ldots,x_K)$.  The ``order'' $n$ assigned to a given term 
in the expansion is the equal to the number of inputs $x_{i_1}$, 
$x_{i_2}$, ..., $x_{i_n}$ involved, also indicated by the labeling of 
weight factors $w_{\lambda,i_1,i_2,\ldots,i_n}$ that appear.  (For 
example $w_{\lambda,4,8,7}$ implies third order.)  In zeroth order, there
just the single bias parameter $w_{0,\lambda}$.  Stopping at the second 
term, of order $n=1$, one has what is known as the Naive Bayes 
classifier, corresponding to the assumption of {\it independent} 
input variables -- hence uncorrelated.  The successive terms 
$n=2, 3, \ldots K$ introduce correlations among two inputs, three 
inputs, etc., up to all $K$ inputs.  The $n$th sum in the expansion 
(\ref{uexp}) contains $K!/(K-n)!n!$ terms.  In practice, this 
expansion is truncated at some maximum order $n_{\rm max} \equiv N$.
As a technical matter, it will be convenient to regard the bias
parameter $w_{\lambda,0}$ as the weight of a connection to unit 
$\lambda$ from a fictitious input $0$ with constant activity 
$x_0 = 1$.

We emphasize that the restrictions imposed on the sums over input indices
$i,j,k,\ldots$ ensure the absence of contributions from ``self-coupling''
weights (or ``self-correlations'') that refer to identical inputs, 
such as $w_{\lambda,ii}x_i^2$.  Accordingly,  the sums run over 
{\it irreducible} correlations among the input variables.  In this 
respect, the HOPP model differs from the most general polynomial 
neural network, while providing a clean separation of correlation 
orders as defined above.

For any given set of weights, the raw output $u_{\lambda}(\textbf{x})$ is 
converted to a normalized probability distribution by processing it
through the following sigmoid ``squashing function'':
\begin{equation}
y_{\lambda}(\textbf{x}) = \frac{e^{u_{\lambda}(\textbf{x})}}
{\sum_{i=1}^{L}e^{u_{i}(\textbf{x})}}.
\label{squash}
\end{equation}
This function has several useful properties, as shown by Clark et 
al.~\cite{clark} for binary inputs and a finite number of outputs $L$.
As already indicated, the architecture of the HOPP model is sufficiently 
general to embody the full statistical correlations inherent in the Bayesian 
approach. A full derivation of this property in the case of binary inputs 
may be found in Ref.~\cite{clark}; an abbreviated version is provided
in the Appendix.   Under the restriction to binary inputs, the model 
simply recapitulates both the structure and content of of Bayes Rule, 
with $y({\bf x})$ being rigorously identified with the Bayesian-optimal 
{\it a posteriori} probability in the limiting sense defined.  Even 
though the proof given does not apply for continuous inputs, the structure 
of the HOPP network is surely general enough to represent with arbitrary 
precision any input-output function of practical interest, although one 
must expect the interpretation of its output as the ideal Bayesian 
{\it a posteriori} probability to be degraded by the deviation from
binary input coding.  

\section{Training Algorithm \& Model Development}

In order to achieve optimal classification, the weights $\textbf{w}$ must 
be adjusted via a training process. The objective is to minimize the 
difference between the probability estimates $y_{\lambda}$ and the actual 
binary values $a_{\lambda}$ known independently for each of a given set 
of input vectors $\textbf{x}$.  This difference is interpreted in
the mean-square sense, so the cost function to be minimized is
\begin{equation}
E = \sum_\lambda (y_\lambda - a_\lambda)^2.
\label{cost}
\end{equation}
In this paper we adopt a gradient-descent method to arrive at the best 
set of weights for the given data set. Two parameters $\epsilon>0$ 
(the learning rate) and $\mu<0$ (called the momentum) and serve to 
control the speed speed and smoothness of descent \cite{haykin}. 
Specifically, the incremental learning rule is taken as
\begin{equation}
\Delta w(t) = \epsilon (a_{\lambda} - y_{\lambda}) 
\frac{\partial u_{\lambda}}{\partial w} + \mu \Delta w(t-1).
\label{lr}
\end{equation}
In this familiar expression, $\Delta w(t)$ is the weight correction 
made in the time step from $t-1$ to $t$, with $w$ as shorthand for the 
complete set of nonzero weights, which have the generic form 
$w_{\lambda, k_{1}k_{2}k_{3}\cdots}x_{k_{1}}x_{k_{2}}x_{k_{3}}\cdots$.

We now describe the steps involved in the implementations of the routine 
implied by Eq.~(\ref{lr}) and developing HOPP models capable of reliable 
generalization.  

It is apparent from Eq.~(\ref{uexp}) that the number $W$ of connection weights 
increases super-exponentially as the order $n$ increases: one is faced
with a potential combinatorial explosion.  To examine this feature
more quantitatively, let us focus on the case $L=2$ and $K=30$ 
that applies for the problem of breast-cancer diagnosis when all input 
features $X_i$ are imposed on the network.  First we note the trivial 
fact that the probabilistic constraint $y_1 +y_2 = 1$ on the output 
signals $y_\lambda$ reduces the number of independent weight parameters 
by half leaving only one free parameter in zeroth order ($n=0$).  By 
virtue of this constraint, one of the two output units becomes redundant, 
as it must necessarily produce the complement of the output of its partner.  
Therefore only a {\it single} output unit $\lambda$ is needed in the 
two-class problem.  In the present application it could indicate 
the probability of malignancy.  Similarly, for the general classification 
problem involving $L$ classes, the HOPP network requires only $L-1$ outputs.

Thus, to zeroth order in the expansion in irreducible correlations
among the inputs, there is only the single bias parameter $w_{\lambda,0}$.
With first-order terms present in Eq.~(\ref{uexp}), we recover the Naive 
Bayes classifier. An additional $K=30$ parameters are introduced,
corresponding to the connection weights appearing in the second term 
of expression (1) for $u_\lambda$.  Already at $n=2$ (leading order in 
{\it correlations} among {\it inputs}), the total number of adjustable 
parameters has climbed to $465 + 1 = 466$, increasing to 4526 at 
$n_{\max}\equiv N=3$, and by $N =4$, the count has come to 31,931.  
Due to the prospect or inevitability of overfitting, it becomes imperative 
to curtail such rapid growth of degrees of freedom.  (For $K=10$, also 
considered in our modeling explorations, the situation is less serious, 
with corresponding parameter counts 11, 56, 176, and 386 at $N=1$, 2, 3, 
and 4, respectively.)  

To deal with this computational complexity, we have adopted the following 
strategy, which has proven both practical and effective. 

\begin{itemize}
\item
Given the database for the problem, consisting of a set of input vectors 
${\bf x}$ with known class assignments, one selects a subset -- the 
training set T -- to be used in training the network model.  

\item
In initializing 
the network, all weights are set to zero, {\it except that} 500 weights 
belonging to orders $0$ to $n_{\rm max}=N$ are given non-zero values drawn 
randomly from a uniform distribution on the interval $[-1.5,1.5]$.

\item
Presenting each of the $P_T$ vectors of this training set in turn, each 
surviving weight is processed through the learning rule (\ref{lr}), 
as the time parameter $t$ is advanced in unit steps. 

\item
Next, the vectors of the training set are permuted randomly among 
themselves, and the process of training continues as before. After 
repeated training cycles for 500 such randomized permutation sets, 
the set of fully evolved weights is culled to yield a chosen maximum of
$W$ surviving weights, priority being given to those of largest
absolute magnitude. (Note: $W$ should be interpreted as the 
maximum number of weights in those special cases where the total
number of adjustable weights is less than $W$, e.g., for the choice
$W =10$.) 

\item
The resulting network model, is then {\it retrained} for 500 permutations 
of the set of training vectors.  The number of post-culling weights are
usually taken to range from $W= 10$ to 50, with 30 as a typical choice.
\end{itemize}

For breast-cancer diagnosis employing the WBCD \cite{dataset} 
these numbers of ``fitting'' parameters are small enough compared with the 
size of the training set that overfitting can normally be avoided. In a 
real-world scenario subserving clinical implementation, the training can 
be an ongoing process, continuing for an indefinite number of generations 
with new data being added to the training pool as they become available.

To provide for validation of the model, the data was arbitrarily divided
into (i) a training set used to generate the model, consisting of 90\%
of the FNAB samples and (ii) a test set comprised of the remainder, 
which is not involved in the determination of weight parameters
and hence the diagnostic model. Performance of the model in 
generalization is judged most intently in terms of the percentage 
of correct classifications for the test set and a measure called the 
Matthews correlation coefficient that removes bias due to imbalance 
in the frequencies of benign and malignant examples in the training set.

To assess the influence of fluctuations depending on which data were 
included in the training set, the whole procedure described above 
was repeated 300 times with different subdivisions of the database 
into training and test sets, thus in the end generating 300 sample 
network models in each of the computational experiments {\bf P1}-{\bf P6} 
described below.  Averages were taken over the performance measures 
computed for each ensemble of sample networks as a basis for overall 
assessment of the quality of the corresponding HOPP classifier.  
Standard deviations from these means were also evaluated, thereby 
achieving an elaborate {\it cross-validation} of the HOPP machine-learning 
approach in its several versions explored here.

Several conventional measures are used to evaluate the performance 
of the HOPP models being developed on both training and test sets.
These are specified in terms of the numbers of true-positive, false-positive, 
true-negative, and false-negative responses of the network for the 
given data set (or subset), denoted respectively by $p_t$, $p_f$, 
$n_t$, and $n_f$. 
\begin{itemize}
\item[$\circ$]
{\it Efficiency} or {\it Accuracy}, defined as the ratio of correct 
classifications to the total number of input patterns $P$ in
the data set presented, thus $ {\rm ACC} =(p_t + n_t)/P$.
\item[$\circ$]
{\it Sensitivity} ${\rm SENS} = p_t/(p_t + n_f)$.
\item[$\circ$]
{\it Specificity}, ${\rm SPEC} = n_t/(n_t + p_f)$.
\item[$\circ$]
{\it Positive Predictive Value} ${\rm PPV}=p_t/(p_t + p_f)$
\item[$\circ$]
{\it Matthews correlation coefficient}, 
\end{itemize}
In addition, there is the {\it Matthews Correlation Coefficient}
\begin{equation}
{\rm MCC} =\frac{p_t n_t + p_f n_f}{\sqrt{p_t + p_f)(p_t + n_f)
(n_t + p_f)(n_t + n_f)}},
\label{MC}
\end{equation}
which compensates for over-representation of one of the outcomes
in the given database.

The outputs of the HOPP network, $y_\lambda$, are actually probabilities
lying in the continuous interval $[0,1]$.  Hence some criterion must be 
imposed to determine the four integers $p_t$, $p_f$, $n_t$, $n_f$  
entering the above performance measures.  The obvious choice is
to set $y_\lambda$ is set equal to 0 if its actual value
is less than 0.5 and to 1 otherwise.  (We note that in clinical practice
this criterion could be modified to reduce the number of false 
negatives.)

\section{Data Set}

The data set used in this study is available in the public domain, compiled 
by Wolberg, Street, and Mangasarian \cite{dataset,wolberg94} from the 
investigations conducted on breast tumors at the University of Wisconsin 
Hospitals, Madison. In the process of Fine Needle Aspiration (FNA), 
a very fine hypodermic needle is inserted into the tumor and a small 
sample of fluid containing tissue and cells is aspirated, mounted 
on a slide and stained to display the nuclei in the cells. The 
pixelated digital images of the cells and their nuclei are then 
subjected to a semi-automatic measurement procedure, aided by a 
computer.  In the image of a given sample, the nuclei are separated 
from the rest of the image, and an approximate boundary is determined 
for each nucleus.  This boundary is treated as a list of points, each 
centered on a pixel.  For the definitions given below, the boundary 
will be represented by the list of vectors $\textbf{b}_{p}$ that 
correspond with the points specifying the nuclear boundary, such that 
the last point, $\textbf{b}_{N}$, and the first point, $\textbf{b}_{1}$, 
coincide (hence there are $N-1$ distinct points).  Measurements are
maded on each of a sample of nuclei from a given biopsy, so as to
obtain values of ten geometrical features $X_i$, as specified below.  
To implement machine learning, or other approaches to diagnosis, 
such a biopsy is represented, in the general case \cite{wolberg95b}, by 
the mean values ${\bar X}_i$ obtained for the variables $X_i$, $i = 1,2, 
\ldots 10$, together with the corresponding standard deviations 
$\sigma_i = [(X_i - {\bar X_i})^2]^{1/2}$ from these means and the 
maximum values $(X_i)_{\rm max}$ obtained for each geometric feature.  In 
addition to the values of these input variables, the full Wisconsin
Breast Cancer data set naturally includes the clinically validated 
classification of the corresponding tumor as either 
benign or malignant.

The ten measured characteristics or features of cell nuclei underpinning
the Wisconsin Breast Cancer Database (WBCD) are:

\begin{itemize}
\item \textbf{Radius} $X_1$.  The center of the nucleus, $\textbf{c}$, is 
defined to be the average of the points on the boundary. The radius 
of the nucleus is defined as the average distance between the points on 
the boundary and the center point, i.e.
\begin{equation}
X_{1} = \frac{\sum_{i=1}^{N-1} \|\textbf{b}_{i} - \textbf{c}\|}{N-1}.
\end{equation}

\item \textbf{Texture} $X_{2}$ The texture of the nucleus is the 
variance of gray-scale intensities in the component pixels inside its
boundary.

\item \textbf{Perimeter} $X_{3}$.  The perimeter of the nucleus is defined as 
the sum of the distances between points as follows:
\begin{equation}
X_{3} = \sum_{i=1}^{N-1} \|\textbf{b}_{i+1} - \textbf{b}_{i}\|.
\end{equation}

\item \textbf{Area} $X_{4}$.  This is defined as the number of pixels
inside the boundary of the nucleus, plus half the 
number of pixels on the boundary.

\item \textbf{Compactness} $X_{5}$.  Observing that the circle has the 
maximum area for a specified value of the perimeter, the compactness 
is defined as the ratio of the square of the perimeter to the area. 
\begin{equation}
X_{5} = \frac{(X_{3})^{2}}{X_{4}}.
\end{equation}
Note that ``compactness'' thus defined actually increases for a more
diffuse nucleus.

\item \textbf{Smoothness} $X_{6}$.  This feature is given by the 
mean absolute deviation of the length $r_{i} = \|\textbf{b}_i - \textbf{c}\|$ 
of a radial line with respect to the radial length of the next adjacent 
neighbor $r_{i+1}$ along the perimeter.  Thus
\begin{equation}
X_{6} = \sum_{i=1}^{N-1} \frac{ |r_{i} - (r_{i} + r_{i+1})/2| }{X_{3}}. 
\end{equation}
It is important to recognize that, as defined, $X_6$ {\it increases} as 
the perimeter becomes more jagged.   Hence it decreases as the perimeter 
becomes more smooth, and so it might better be called Roughness.

\item {\bf Concavity} $X_{7}$.  
This feature represents the severity of indentations on the periphery 
of the nucleus.  To quantify it, a simple algorithm is used to construct 
chords between points on the perimeter so as to bound such indentations 
and measure the total area of the concavities so enclosed. Since the chord 
lengths involved are small, this feature quantifies small-scale rather 
than large-scale irregularities \cite{wolberg93a,wolberg93b,wolberg94}.

\item {\bf Concave Points} $X_{8}$.  This feature is similar 
to concavity in that it measures the number of representative points on the 
perimeter that lie within the concave regions of the image. 
In contrast to the $X_7$, the magnitudes of the deviations from regularity 
affect this measure.  It is to be noted that the measures $X_7$ and $X_8$ 
quantifying concavity and concave points are affected by pixelation. 

\item {\bf Symmetry} $X_{9}$.  This feature actually represents the 
{\it deviations} from symmetry in the direction transverse to the
major axis (defined as the longest chord passing through the center). 
Lines perpendicular to the major axis are drawn at regular intervals 
and the lengths of the intercepts, $l_{ai}$ and $l_{bi}$ are used to define
\begin{equation}
X_{9} = \frac{\sum_{i} |l_{ai} - l_{bi}|}{\sum_{i} |l_{ai} + l_{bi}|}.
\end{equation}

\item {\bf Fractal Dimension} $X_{10}$.  Conceptually, this measure has 
been introduced by Mandelbrot \cite{mandelbrot}, in particular to describe 
an irregular boundary of two-dimensional area, such as the coastline of 
Norway. According to Mandelbrot's prescription, the fractal dimension 
of the boundary of a cell nucleus is estimated by measuring the perimeter 
of its nucleus with increasingly larger ``rulers.'' The log of the 
perimeter measurements so obtained is plotted against the logarithm
of the ruler size.  In the limit, the magnitude of the slope of the
plot gives the fractal dimension $X_{10}$ of the nuclear perimeter [3].

\end{itemize}

Figs.~3-7 show histograms of the data extracted for the ten nuclear 
features $X_1$ through $X_{10}$ as extracted from measurements 
performed on the FNA biopsies of all 569 patients.  These represent
approximate probability density distributions of the geometrical
properties that have been singled out.  Note the broad distributions 
and the absence of multiple peaks.  A single sample would clearly
present a challenge for manual diagnosis.

The fact that these figures show the distributions for all biopsies, 
both well and ill, suggests that one might also consider separately 
the distributions for benign and malignant cases, and the degrees 
of overlap for each feature.  This could offer another route to 
diagnosis, which might even be extended beyond the binary 
benign-malignant decision to a measure of the degree of illness.

\begin{figure}
\begin{center}
\includegraphics[width=.48\textwidth]{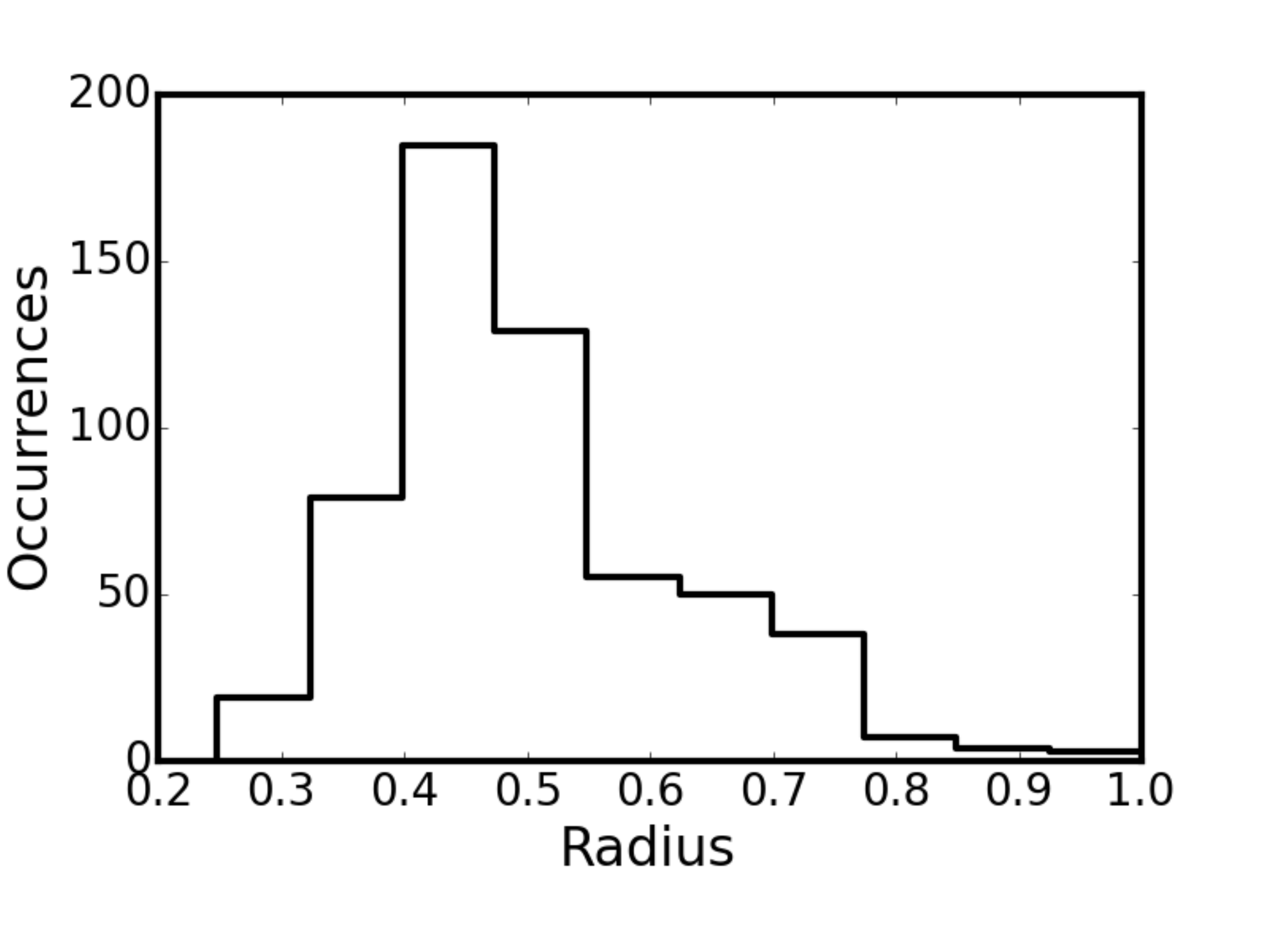}
\includegraphics[width=.48\textwidth]{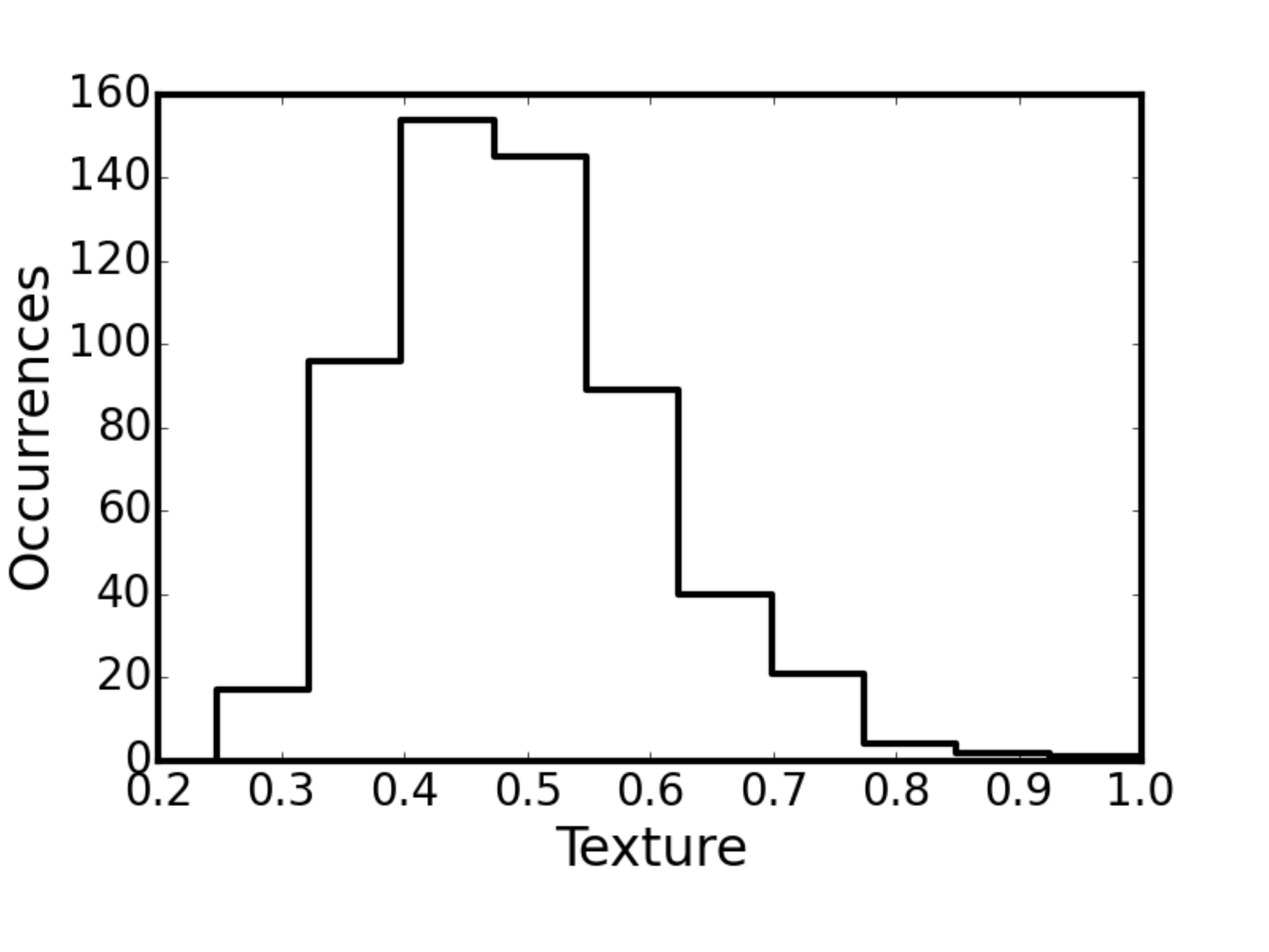}
\end{center}
\caption{Frequencies of occurrence (approximate probability 
densities) of different values of input features $X_1$ (left panel)
and $X_2$ (right panel), normalized as specified in the
text and plotted as normalized distributions among bins of
each measured variable.  These features correspond respectively to 
the sample means of radius and texture metrics
as determined in FNA biopsies of the 569 patients of the Wisconsin 
data set.} 
\label{distribs1}
\end{figure}

\begin{figure}
\begin{center}
\includegraphics[width=.48\textwidth]{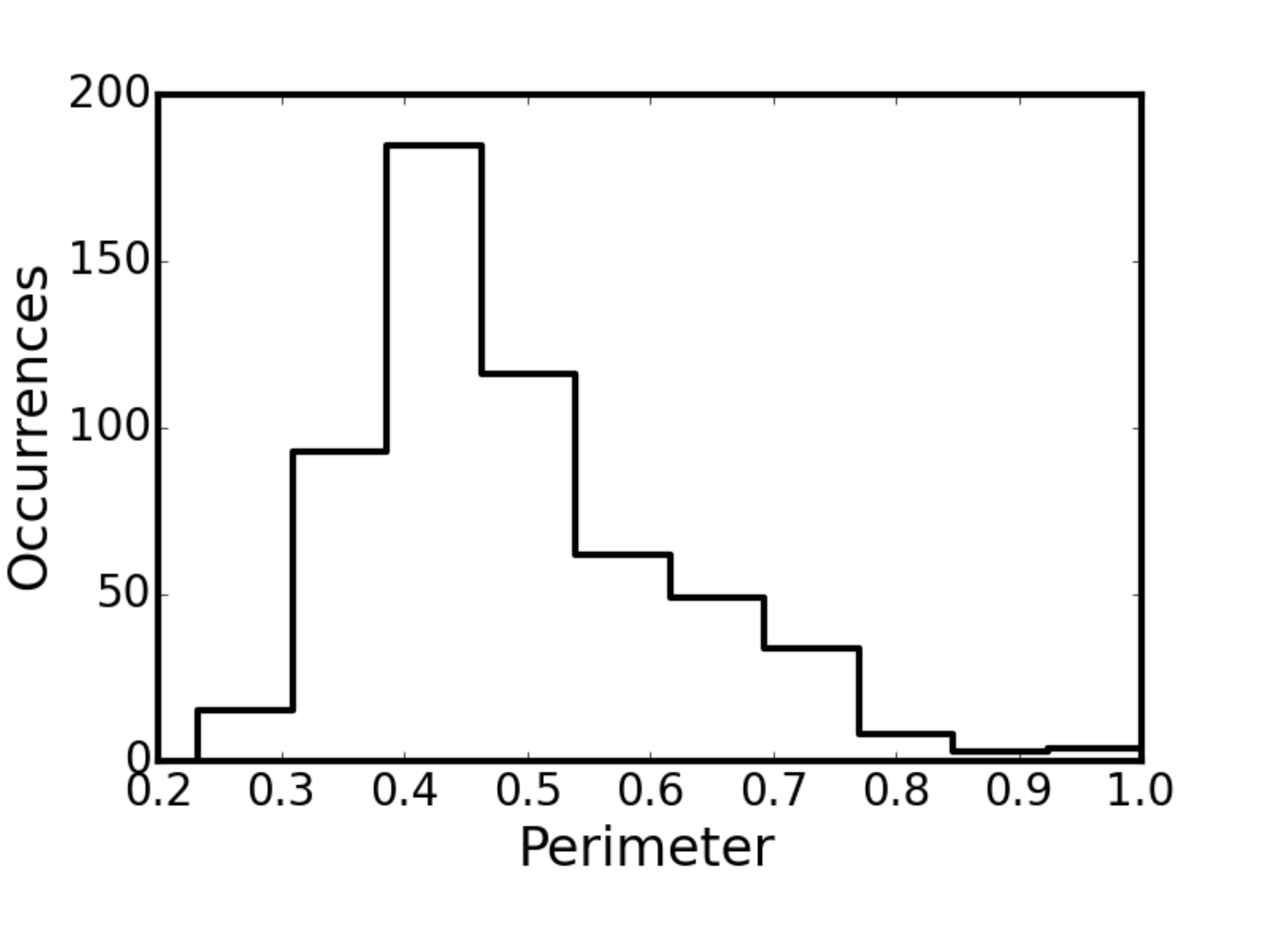}
\includegraphics[width=.48\textwidth]{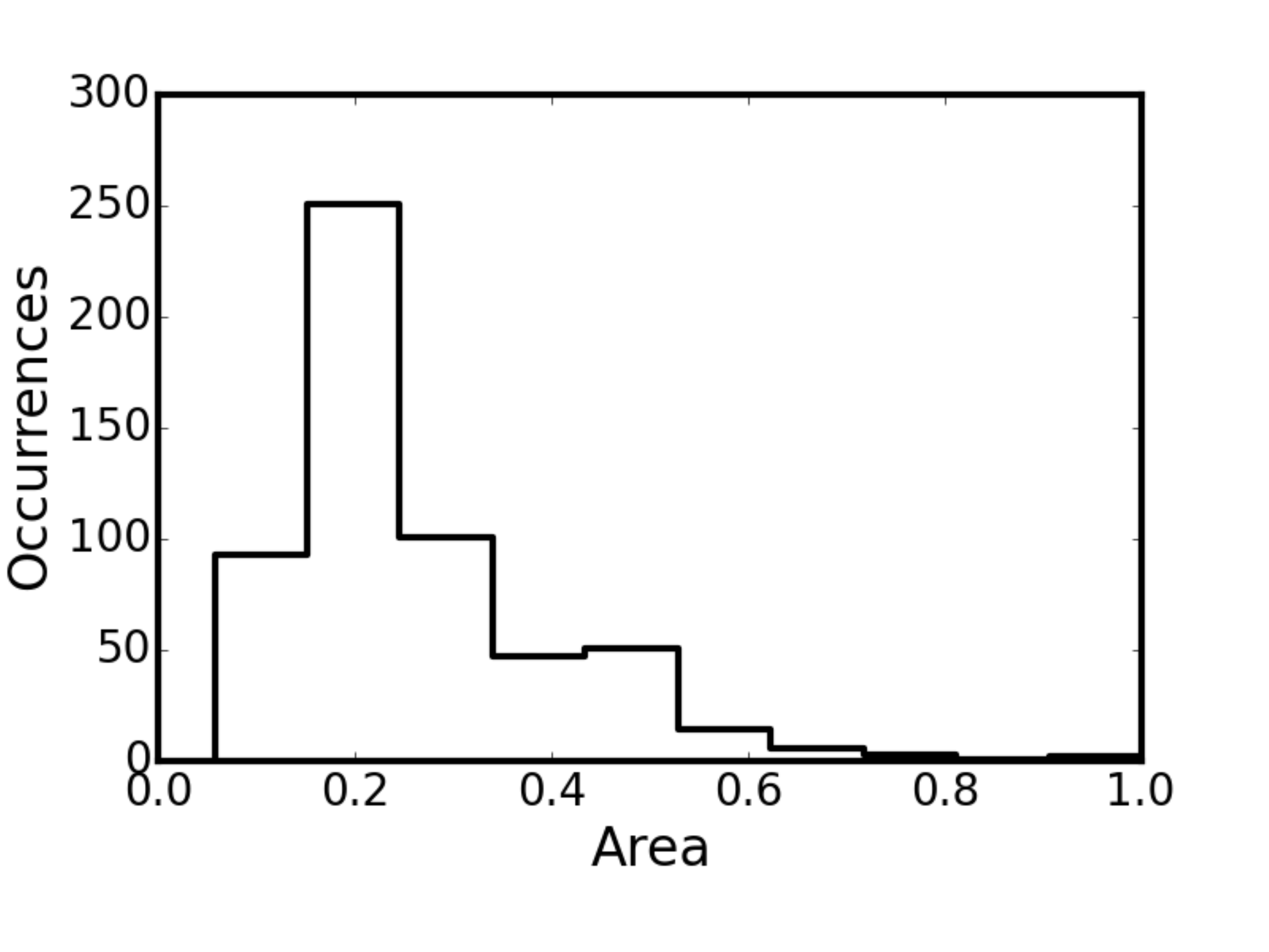}
\end{center}
\caption{As in Fig.~2 but for input features $X_3$ and $X_4$
corresponding respectively to sample means of perimeter and
area metrics.} 
\label{distribs2}
\end{figure}

\begin{figure}
\begin{center}
\includegraphics[width=.48\textwidth]{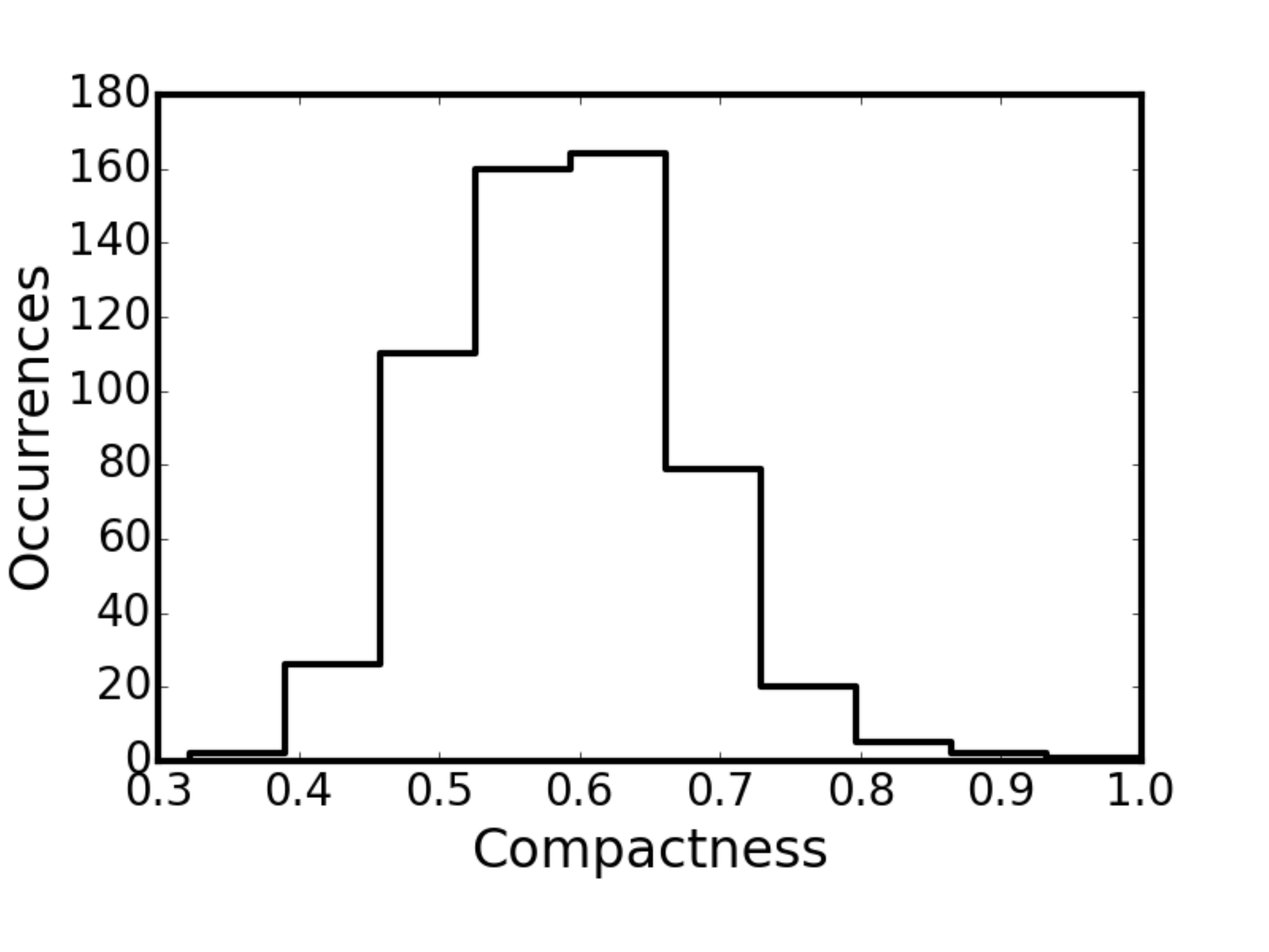}
\includegraphics[width=.48\textwidth]{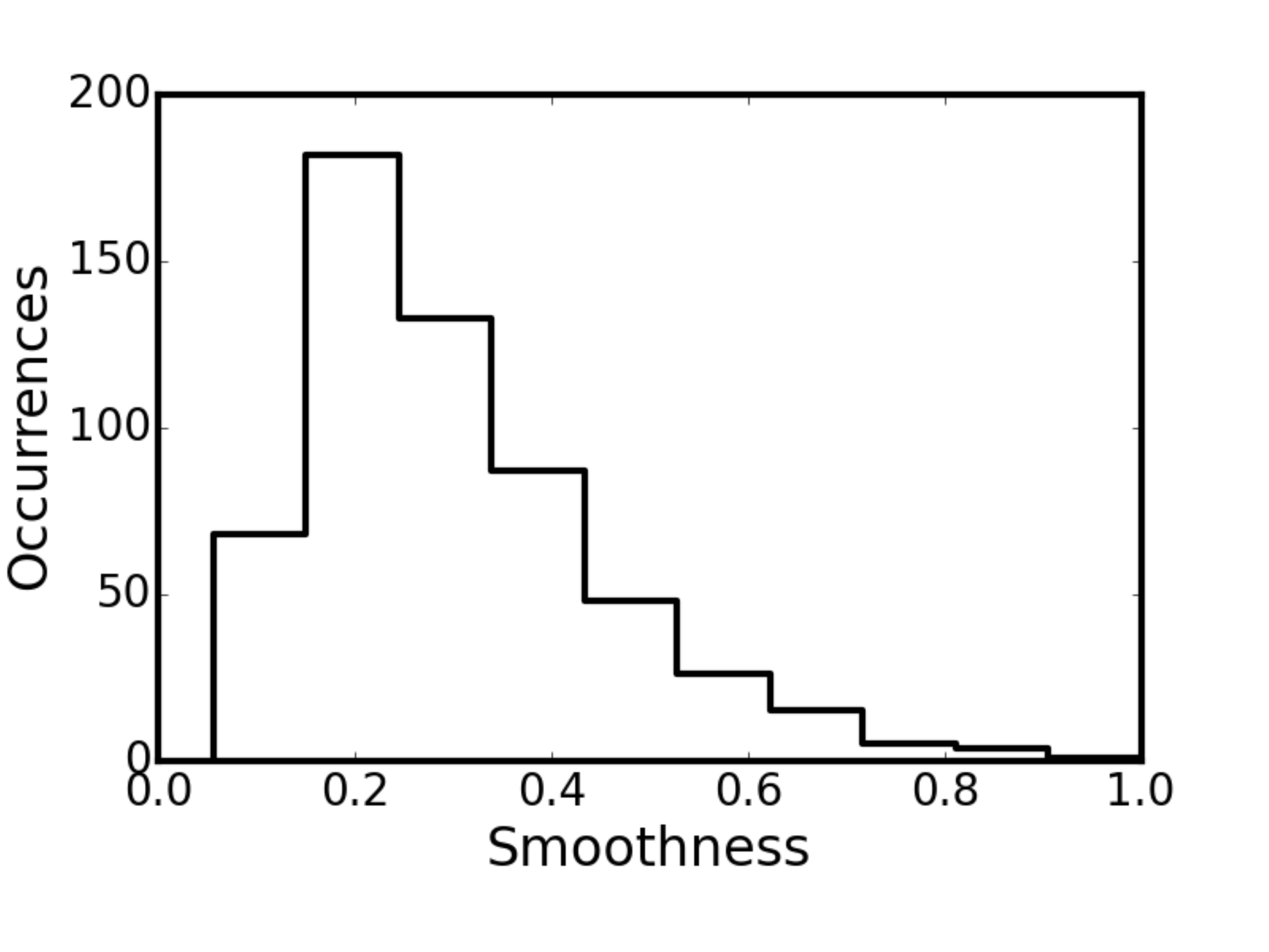}
\end{center}
\caption{As in Fig.~2 but for input features $X_5$ and $X_6$
corresponding respectively to sample means of compactness and
smoothness metrics.} 
\label{distribs3}
\end{figure}

\begin{figure}
\begin{center}
\includegraphics[width=.48\textwidth]{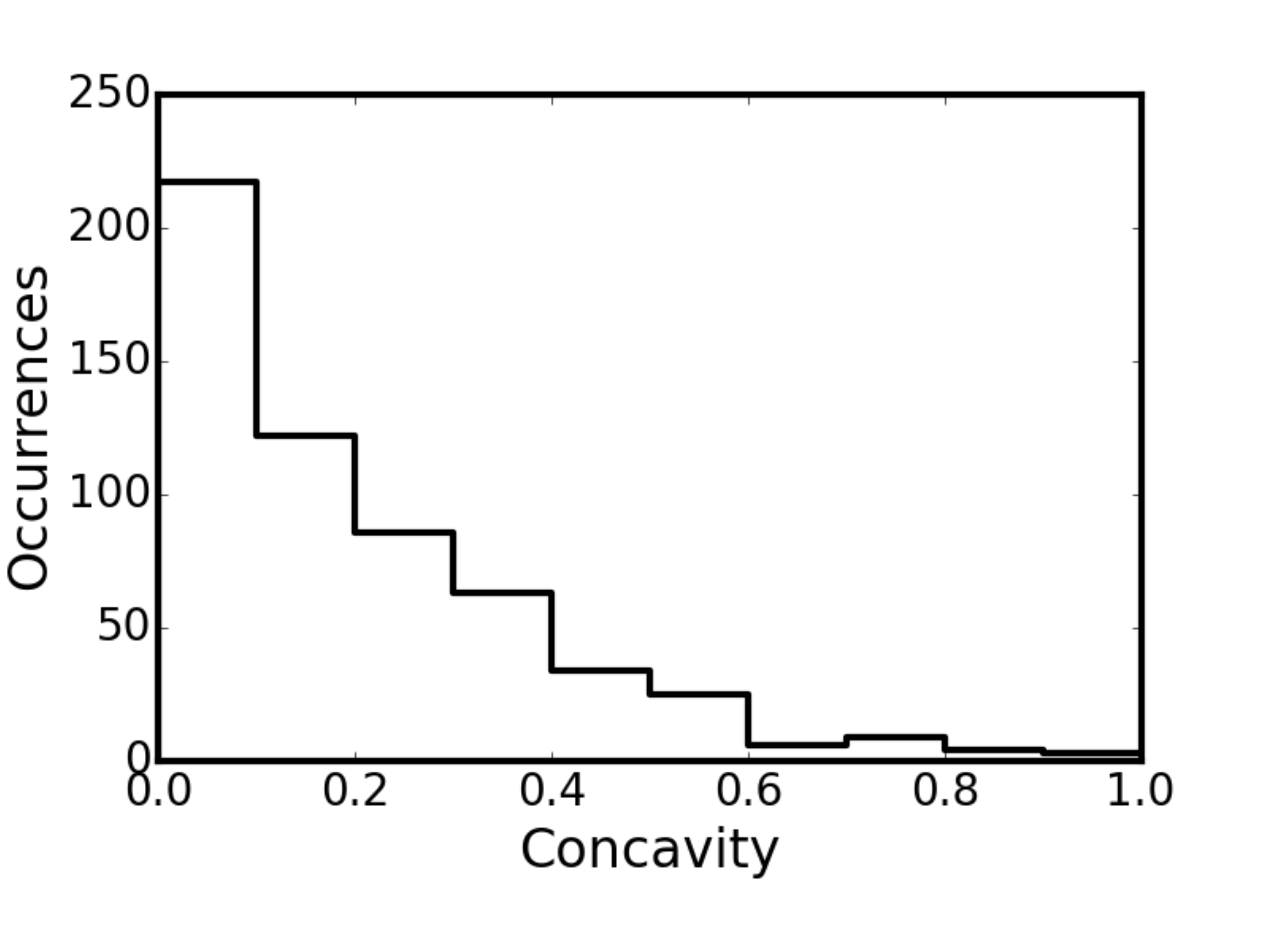}
\includegraphics[width=.48\textwidth]{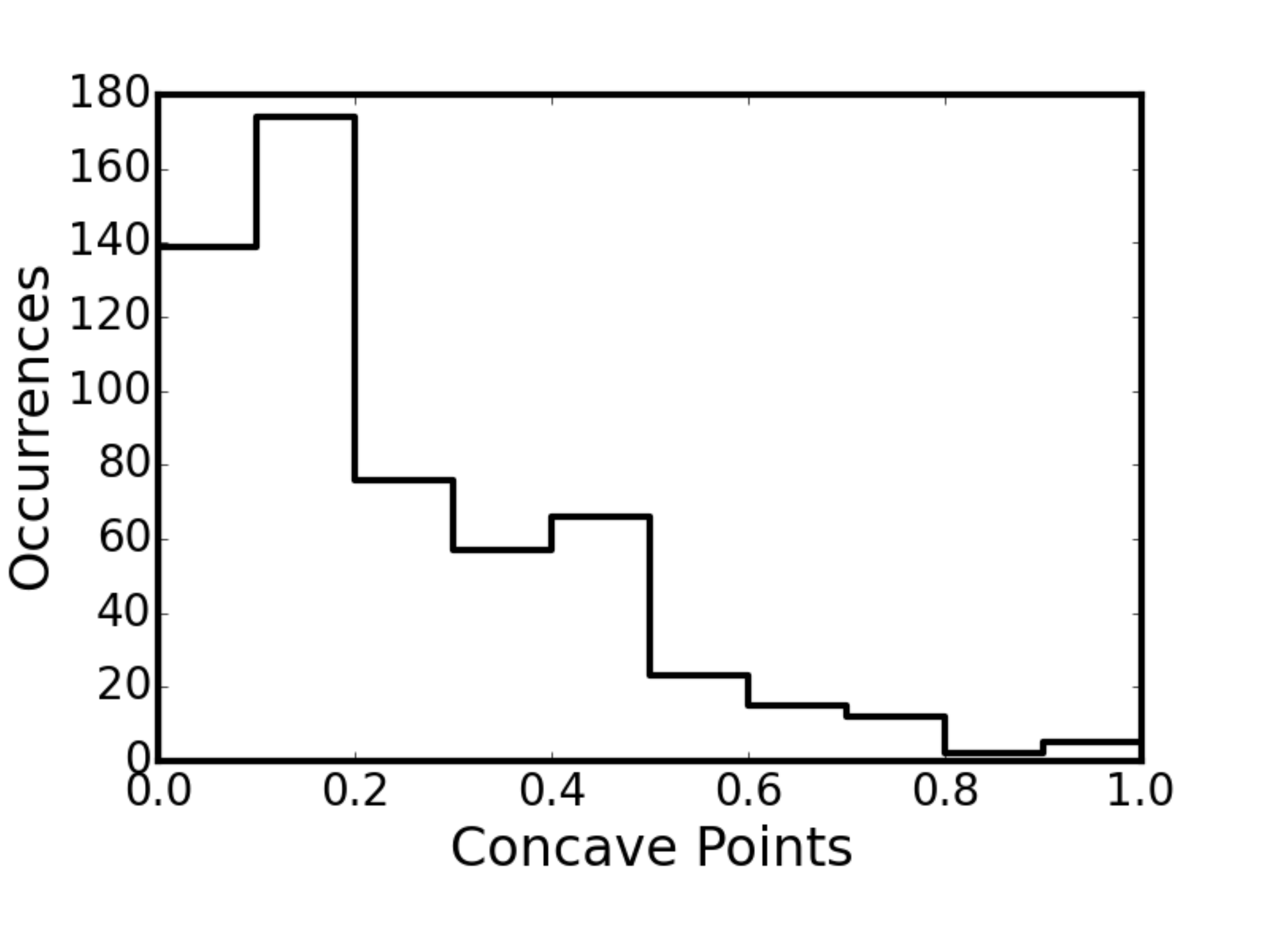}
\end{center}
\caption{As in Fig.~2 but for input features $X_7$ and $X_8$
corresponding respectively to sample means of concavity and
concave points metrics.} 
\label{distribs4}
\end{figure}

\begin{figure}
\begin{center}
\includegraphics[width=.48\textwidth]{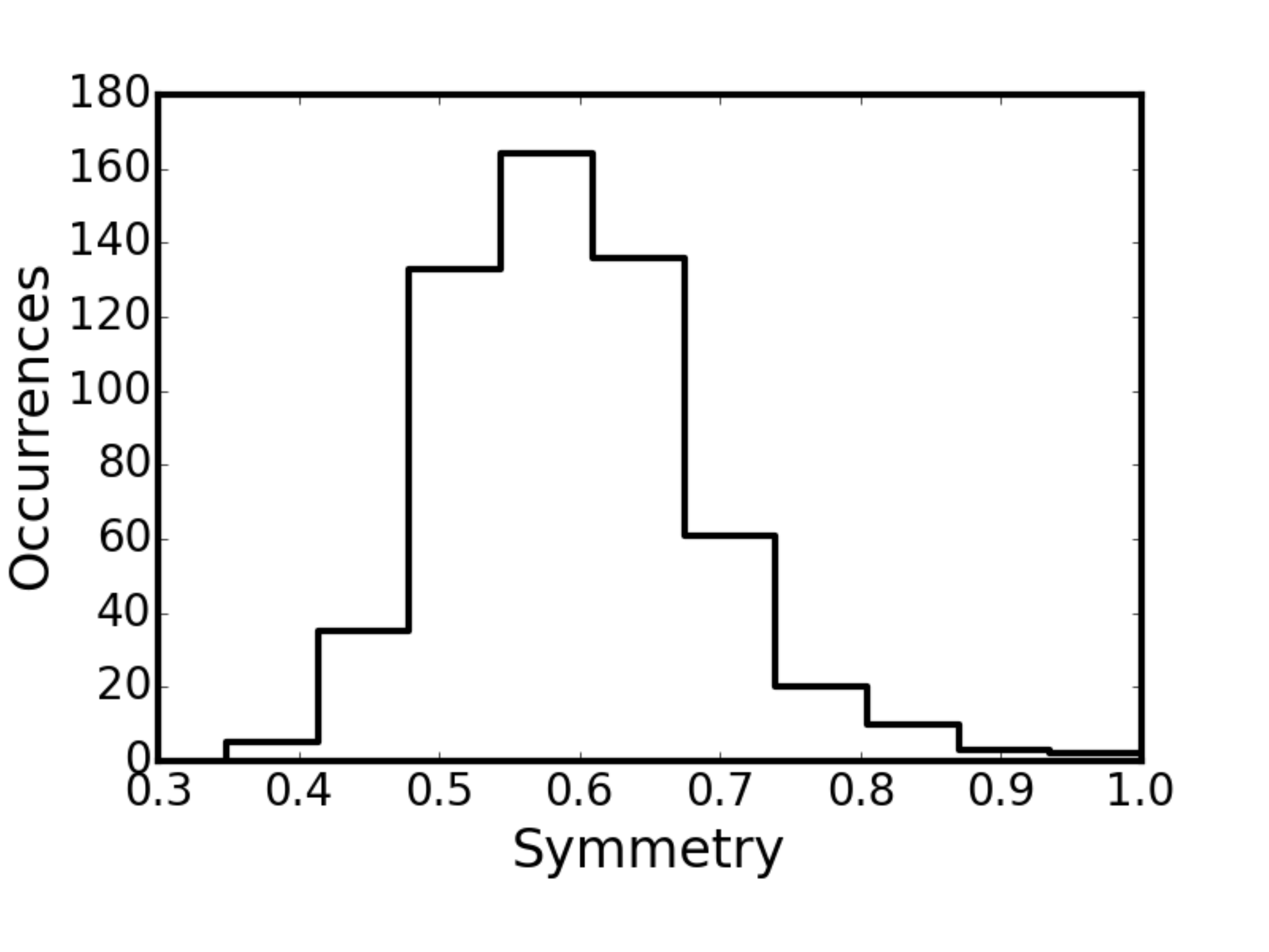}
\includegraphics[width=.48\textwidth]{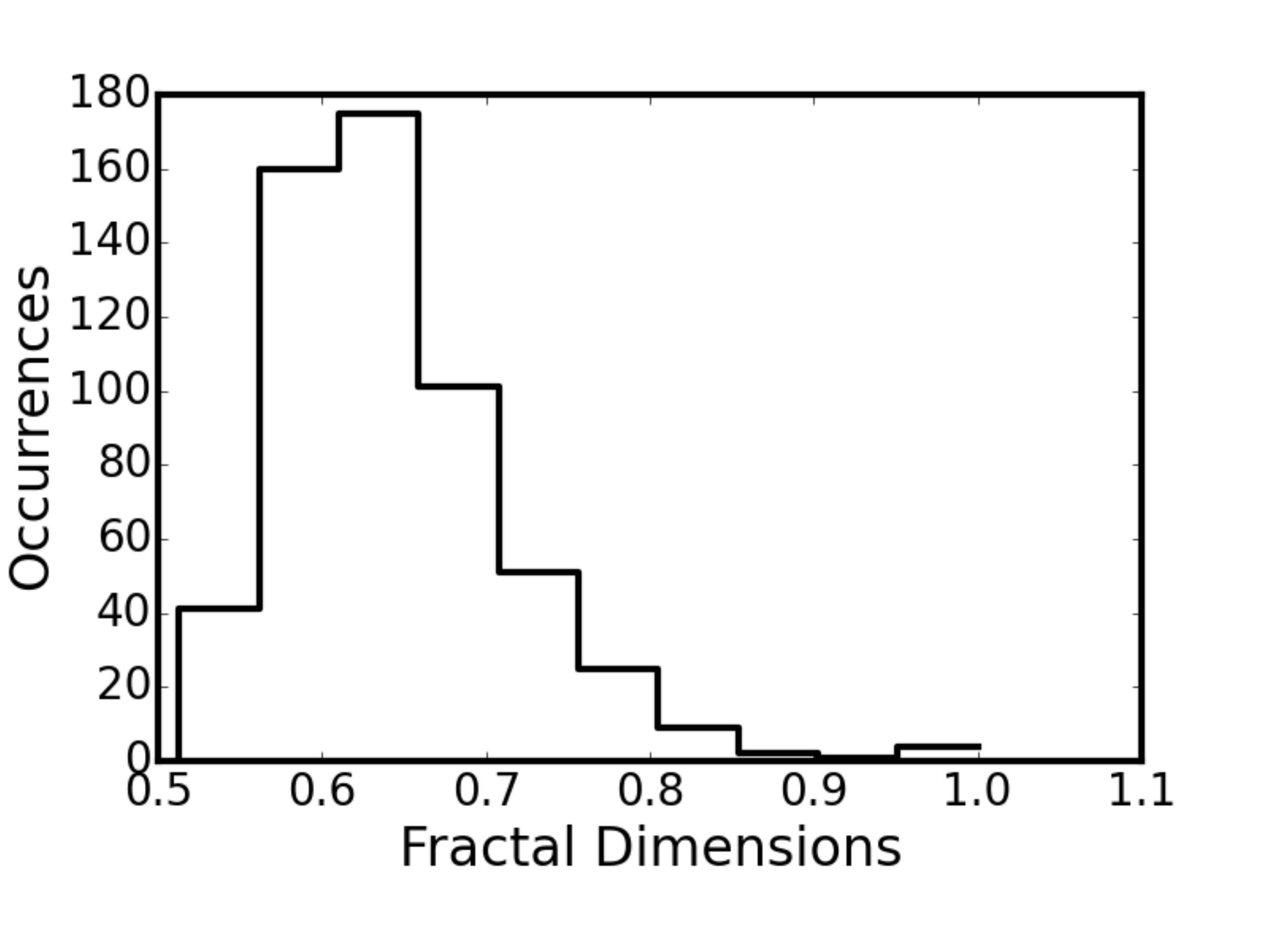}
\end{center}
\caption{As in Fig.~2 but for input features $X_9$ and $X_{10}$
corresponding respectively to sample means of symmetry and
fractal dimension metrics.} 
\label{distribs5}
\end{figure}

\section{Strategies Explored \& Model Performance Evaluated}

The results reported here are based on several distinct training
strategies for the construction of HOPP classifiers, each with its 
own objective.  The quality of the resulting network models 
is evaluated in terms of the performance measures listed in Section 
5, namely: accuracy ACC (or efficiency), sensitivity SENS, 
specificity SPEC, positive predictive power PPV (or precision), and 
the Matthews correlation coefficient MCC defined by Eq.~(\ref{MC}).  
Since we are primarily concerned with performance in prediction
or ``generalization,'' we focus on the measures ACC and MC obtained
for the test sets, but some results are also also quoted for 
the other three measures.  Of course, one is also interested the
quality of the model in reproducing known cases, i.e., ``quality
of fit,'' which is reflected in the results for the five performance 
measures on the training sets.

It is well to note here that the above performance measures refer
to strict yes-or-no decisions, whereas the HOPP network model 
automatically yields a {\it probability estimate} for malignancy 
and its complementary estimate for a benign condition.  This actually 
presents no problem, in that the lack of ambiguity is ensured by the 
strongly bivalent character of the network output, as clearly
documented in our results.

We present here a summary what has been learned from a large number 
of computer experiments carried out according to eight alternative 
training protocols.  The reader has access to a set of tables 
providing detailed results for all eight protocols at 
\url{http/:physics.wustl.edu/people/john-w-clark}.

\begin{itemize}
\item {\bf Procedure 1 (P1)} In our initial approach to model 
development, all of the $K=30$ features $X_1,\ldots,X_{30}$ determined for 
the FNA biopsies were used in constructing the neural-network classifiers.  
Weights of order $n=3$ or lower were included among those sampled, i.e.,
$N = 3$, so weights as complicated as $w_{\lambda,ijk}$ are included.
\end{itemize}

\begin{itemize}
\item[(i)] The HOPP networks constructed are able to classify 95.70\%
of the 300 {\it test sets} correctly {\it on average}, i.e., 
with an average accuracy ${\rm ACC} = 0.9570$ on the model ensemble, 
with standard deviation 0.0273.  The corresponding result for the 
Matthews coefficient MCC is $0.9077 \pm 0.0589$, and for the other
performance measures we find ${\rm SENS} =  0.9767 \pm 0.0256$, 
${\rm SPEC} = 0.9234 \pm 0.617$ and ${\rm PPV} =  0.9559 \pm 00351$.
The quality measures achieved on the {\it training sets} in this computer
experiment are ${\rm ACC} = 0.9703 \pm 0.0053 $ and ${\rm MCC} 
= 0.9364 \pm 0.0114$, ${\rm SENS} = 0.9836 \pm 0.0052$, ${\rm SPEC} 
= 0.9480 \pm 0.0083$, and ${\rm PPV} = 0.9696 \pm 0.0048$.
Such performance was realized with $W = 30$ post-culling weights (plus
the two biases).  
Such performance was realized with $W = 30$ post-culling weights (plus
the two biases).  It is likely that a slight improvement $\sim 1-2\%$ 
can be gained by adjusting this choice.
Some of the sample models among the 300 involved in cross-validation 
process demonstrated perfect classification.  

\item[(iii)] As described in Section 4, a thorough cross-validation 
process was implemented by the creation of 300 HOPP models for 300 
different randomly generated subdivisions of the database into 
training and test sets.  The results summarized in (i) provide 
an unambiguous demonstration of the reliability of the HOPP classifier, 
with additional evidence to be provided in further numerical experiments.  
For all five performance measures, the standard deviations from the 
corresponding means are very small, at most a few percent of the means.

\item {\bf Procedure 2 (P2)}.  To provide a baseline case that has 
been studied in previous machine-learning studies of this classification
problem, we revert to the Naive Bayes classifier, thus stopping at first 
order in irreducible correlations, ignoring the higher-order interactions 
in Eq.~(\ref{uexp}). Thus we truncate the expansion at $N = 1$, thereby 
allowing for weights of the form $w_{\lambda,i}$ (along with the biases  
$b_\lambda$).  In a first modeling exercise, the number weights 
was culled to a maximum $W$ of 30.  In this case, the 300 network models 
that were created during cross validation attained an ensemble-average 
ACC of 0.966 and mean Matthews coefficient MCC of 0.928 on the test 
sets, hence even slightly better predictive performance than that found 
to third order by procedure {\bf P1}, also for 30 surviving non-trivial 
weights.  With 50 remaining weights, the average performance figures 
on the test sets improved to ACC = 0.9725 and MCC = 0.9411, the 
corresponding results for the training set being 0.9866 and 0.9714, 
respectively.  One must conclude that with all 30 available inputs, 
the Naive Bayes classifier by itself can already display splendid 
performance.  The quantitive effectiveness of this simple algorithm
for the Wisconsin database has been documented previously in 
Refs.~\cite{karabatak2,chaurasia} (In this special classification 
problem, it seems that ``we can put all our eggs in one basket.'')

\item {\bf Procedure 3 (P3)}.  In a third model-development exercise, 
the set of input features used in training was restricted to
the {\it mean values} ${\bar X}_i$ of the ten measured quantities 
$X_1,\ldots,X_{10}$.  One motivation of this experiment was to test the 
sensitivity of predictive performance of the HOPP model to a reduction
in the number of input parameters used to characterize the space in
which the diagnostic problem is modeled.  Again 300 network models were 
developed in cross-validation runs, now with the number of post-culling 
weights $W$ ranging from 10 to 40 in steps of 10.  Thus, four sets of 
cross-validation runs were made for each choice of the maximum order 
$N$ retained in the correlation expansion (\ref{uexp}), which 
ranged from 1 (Naive Bayes) to 4.  In this case, the best test-set 
results at the Naive Bayes level, obtained at $W=20$, are ${\rm ACC} = 
0.9373 \pm 0.0327$ and ${\rm MCC} = 0.8656 \pm 0.0701$, along with  
${\rm SENS} = 0.9572 \pm 0.0378$, ${\rm SPEC} = 0.9040 \pm 0.0647$,
and ${\rm PPV} = 0.9440 \pm 0.0377$.  Overall, the {\it optimal} results
on the test ensemble are found at second order, $N = 2$,
and $W=20$: ${\rm ACC} = 0.9427 \pm 0.0292$ and ${\rm MCC} = 0.8770 
\pm 0.0620$, with  ${\rm SENS} = 0.9606 \pm 0.0342$, ${\rm SPEC} = 
0.9121 \pm 0.0625$, and ${\rm PPV} = 0.9497 \pm 0.0364$. 

The choice of ``best'' model is somewhat ambiguous, depending on what
aspect of the diagnosis is considered most important.  For example,
a physician my give priority to suppressing false negatives for presence
of the disease and would then be most interested in the sensitivity
of the diagnostic tool.  Be that as it may, in identifying the optimal
model among those generated in a given Procedure, we shall instead 
give priority to the accuracy or efficiency of prediction, ACC, 
followed by the Matthews correlation coefficient MCC. It will be 
seen, however, that the differences in the quality of the results 
obtained over the ranges of $N$ and $W$ considered are generally 
minimal; basically, the differences are in the noise.  This applies
to the documented performance on both test- and training-set 
ensembles.  

With regard to the training phase, performance refers to the quality
of fit, which generally improves with increasing $N$ and $W$; 
however, the gradual increases in the five quality indices remain below 
$3\%$ over the ranges considered, and overfitting does not appear to be an 
issue. In the {\bf P3} study, the training-set results at $N = 2$ and 
$W=20$ are ${\rm ACC} = 0.9497 \pm 0.0049$ and ${\rm MCC} = 0.8924 \pm 
0.0102$, with ${\rm SENS} = 0.9839 \pm 0.0215$, ${\rm SPEC} = 0.9661$, 
and ${\rm PPV} = 0.9798 \pm 0.0231$.   

Summarizing,
\begin{itemize}
\item[(i)]
For the 10-input case based only on the mean values of the ten
$X_i$, we find only a modest degradation in the measure 
${\rm ACC}$ from the results found with Procedure 1.
\item[(ii)]
Reinforcing what was learned in comparing the results of exersises
{\bf P1} and {\bf P2} for the 30-input case, we again find that the Naive
Bayes approximation yields results not significantly worse than
the best results obtained at higher order (in this case $N=2)$, 
and with the same number of fitting parameters.
\end{itemize}

\item {\bf Procedure 3a (P3a)}. 
A clear message from exercises {\bf P1}-{\bf P3} is that for the diagnostic
problem defined by the WBCD, higher-order irreducible correlations 
(i.e., terms with $n>1$ in the expansion (\ref{uexp}) are in fact 
of minor importance.  That being so, it might be advisable to give 
these these higher-order effects less emphasis in the weight-culling 
process.  With this in mind, we have repeated Procedure 3, but in 
deciding upon the weights to be kept, we used the $n$th-roots of the 
magnitudes of the $n$th-order weights, rather than the weights 
themselves, in establishing priority for survival. 

The procedure so executed, otherwise identical to {\bf P3} and labeled 
{\bf P3a}, yielded the next-best results among all the computational 
procedures explored for HOPP models, if the maximum number of weights
kept is restricted to $W=30$.  The optimal results on the 
test-set ensemble, occurring rather naturally at the Naive Bayes 
stage and for the choice $W = 30$, are ${\rm ACC} = 0.9727 \pm 0.0196$ 
and ${\rm MCC} = 0.9421 \pm 0.0418$.  The improvement over results 
from {\bf P1} and {\bf P2} may be only marginally significant for 
${\rm ACC}$, but the improvement of the Matthews coefficient is quite 
striking.  (In general, the latter improvement is also found for the 
other results obtained with this procedure.) 

\item {\bf Procedure 3b (P3b)}.  Complementary to {\bf P3} and {\bf P3a}, 
we have repeated {\bf P3} for the ten inputs provided by the so-called
``worst'' values of the quantities $X_1,...,X_{10}$ measured for
cell nuclei in a given biopsy, as defined in Section 5.  Significantly 
better results are obtained than were found for {\bf P3}.  In fact, the 
quality of the results obtained using this alternative to {\bf P3} is the 
best achieved for any of the eight HOPP-based procedures applied to this 
classification problem, although {\bf P3a} yields competitive results.  
However, in this case the optimal test-set performance was found at
$W=40$ weights: ${\rm ACC} = 0.9772 \pm 0.0196$, ${\rm MCC} = 0.9511 
\pm 0.0420$, ${\rm SENS} = 0.9839 \pm 0.0215$, ${\rm SPEC} = 0.9661$, and 
${\rm PPV} = 0.9798 \pm 0.0231$.   Significantly, the corresponding 
Naive Bayes results for $W=30$, and even those for $W=20$, are very close 
to these, notably ${\rm ACC}_{W=30} = 0.9770 \pm 0.0176$, ${\rm ACC}_{W=20} = 
0.9769 \pm 0.0190$, ${\rm MCC}_{W=30} = 0.9500 \pm 0.0394$, and 
${\rm MCC}_{W=20} = 0.9505 \pm 0.0411$.  Moreover, the best results
(versus $W$) for the higher-order cases $N > 1$ show only slight 
degradation, typically under $1\%$.  Given the relatively small numbers 
of adjustable parameters in these networks, one normally expects the 
ensemble averages of the five performance measures obtained for the training 
ensembles to fall short of perfection on average.  Nevertheless, they 
exhibit remarkably high quality.  As a typical example, in the Naive Bayes 
case we find ${\rm ACC}_{W=30} = 0.9825 \pm 0.0028$ and ${\rm MCC}_{W=3} 
= 0.9625 \pm 0.0060$.

\item {\bf Procedure 4 (P4)}.  It is of considerable interest to develop 
models for a {\it reduced set} of inputs that are are judged, by some 
plausible criterion, to be especially most influential in successful 
classification.  With fewer inputs, hence fewer adjustable parameters 
at given correlation order, one has the concomitant advantage of 
suppressing overfitting.  This general theme was already fruitfully 
explored in the pathbreaking work of Wolberg and collaborators 
\cite{wolberg93b,wolberg94}.  They were able to identify small subsets 
of the standard 30 input features, which, as inputs for lower-dimensional 
multisurface classifiers \cite{mang68,mang93,bennett92a,bennett92b}, 
maintained accurate representations of the training data while providing 
for good generalization to new cases.  In particular, it was found that 
the best single-plane diagnostic classifier based on three features, 
namely mean Texture, worst (maximum) Area, and worst Smoothness, was 
able to classify 97.3\% of the cases in the database correctly. The 
performance on cases {\it not} used in model construction was assessed 
by ten-fold cross-validation, with a predicted success rate of 97.0\% on the 
``unseen'' examples, as appraised through a ten-fold cross-validation.
Interestingly, Street et al.\ \cite{wolberg93b} point out that the
best feature triplets tend to contain a size feature, a shape feature,
and a third feature of indifferent character with marginal relevance.

In procedure {\bf P4} we have generated ensembles of HOPP models 
based on the same three input features employed by Wolberg et al.\ in 
the study cited above.  Otherwise, this exercise follows essentially 
the same pattern as {\bf P1}-{\bf P3b}.  We considered values of 
$W$ ranging from 10 to 80 in steps of 10, for maximum orders
$N$ up to 3.  The best results for the test-set ensembles were 
obtained at (i) $N=1$ with $W=80$, (ii) $N=3$ with $W=70$, and 
(iii) $N=3$ with $W=80$, yielding (i) ${\rm ACC} = 0.9570 \pm 0.0263$, 
${\rm MCC} = 0.9091 \pm 0.0555$, (ii) ${\rm ACC} = 0.9570 \pm 0.0259$, 
${\rm MCC} =  0.9087 \pm 0.0542$, and (iii) $0.9573 \pm 0.0243$, 
${\rm MCC} = 0.9091 \pm 0.0516$.  The associated training-set results 
have respectively (i) ${\rm ACC} = 0.9565 \pm 0.0034$, ${\rm MCC} = 
0.9068 \pm 0.0072$, (ii) ${\rm ACC} = 0.9606 \pm 0.0037$, 
${\rm MCC} = 0.9157 \pm 0.0078$, and (iii) ${\rm ACC} = 0.9601 \pm 0.0036$, 
${\rm MCC} = 0.9146 \pm 0.0075.$  We find as before that, within
the statistical noise, the Naive Bayes classifier offers performance
on a par with the best models including explicit effects of higher-order 
correlations.  The other three test-set performance measures for the 
best Naive Bayes model ensemble are ${\rm SENS} = 0.9681 \pm 0.0319$, 
${\rm SPEC} = 0.9399 \pm 0.0513$, and  ${\rm PPV} = 0.9632 \pm 0.0319$, 
while the corresponding training-set results are ${\rm SENS} = 0.9692
\pm 0.0061$,  ${\rm SPEC} = 0.9350 \pm 0.0105$, and 
${\rm PPV} = 0.9618 \pm 0.0059$.

\item {\bf Procedure 5 (P5)}. 
Consider once again the 30 individually measured features of the cell 
nuclei of a given FNAB, proposed by Wolberg et al.\ \cite{wolberg90} as 
potentially decisive evidence for reliable diagnosis, and alternatively
a smaller set of 10 features consisting for example of the mean values
or maximum values of the features $X_1,\ldots,X_{10}$ identified in Section
5.  In either case the terms of expansion (\ref{uexp}) beyond the Naive Bayes 
order $n=1$ describe the irreducible effects of these features acting in 
unison, i.e., of {\it factors} composed of the original features.  For
30 features, the same combinatoric analysis as presented in Section 4 
shows that the 2nd-, 3rd-, and 4th-order correlations in this expansion give 
rise respectively to 466, 4526, and 31,931 {\it distinct} factors,  
the respective counts being reduced to 46, 176, and 386 if only
10 features are employed.  In a transparent notation, simple
examples of such factors are: Area$\times$Smoothness ($n=2$), 
Perimeter$\times$Concavity$\times$Symmetry ($n=2$), and 
Radius$\times$Texture$\times$Compactness$\times$FractalDimension 
($n=4$).  (For each example a choice is to be made among mean, 
worst-value, and standard deviation of the each property involved.)

In some classification problems, the effects of features acting {\it in 
combination} (i.e., nontrivial ``factors'') may be stronger or weaker
than the same features acting in isolation, i.e., there may be terms 
in Eq.~(\ref{uexp} beyond the Naive Bayes approximation that 
play a special or even dominant role in determining the class of a 
given example.  Execution of Procedures {\bf P1}-{\bf P4} has provided no
evidence that this behavior is seriously in play for the classification 
task posed by the WBCD.  However, in other classification problems, 
the following criterion might prove useful in assessing the degree 
of relevance of a given factor (or feature, as a special case) 
for successful generalization.  Let the {\it number of appearances} 
$A$ (or frequency) of a given factor in any of a set of model 
development runs (cross-validations) be defined as the number of 
times that particular factor survives in the culling process leading 
to an ensemble of HOPP networks.  One may then rank order the 
factors according to the counts $A$ for this ensemble, recognizing 
that there will be many degeneracies.  

Appearance counts $A$ were made in some of model-development runs described
above.  As expected, there was high degeneracy among the counts for
various factors and individual features, there were many that
appeared in all or nearly all of the 300 fully processed networks. 
among the 300 .  In an exercise that mirrors that reported
in Refs.~\cite{wolberg93b,wolberg94}, we have created an ensemble 
of 300 HOPP networks having mean values of Texture, Area, and Concave 
Points, which scored at or close to 300 in appearances.  
For these runs we considered the same ranges of $N$ and $W$ as in 
{\bf P4}.  The best results from this set of runs were found at 
$W=10$ for $N = 2$.  For this ensemble, the test-set results are 
${\rm ACC} = 0.9414 \pm 0.0306$, ${\rm MCC} = 0.8741 \pm 0.0659$, 
${\rm SENS} = 0.9629 \pm 0.0334$, ${\rm SPEC} = 0.9056 \pm 0.0690$, 
and ${\rm PPV} = 0.9460 \pm 0.0398$, the associated training-set
values being ${\rm ACC} = 0.9409 \pm 0.0049$, ${\rm MCC} = 0.8732 \pm 0.0102$, 
${\rm SENS} = 0.9624 \pm 0.0107$, ${\rm SPEC} = 0.9045 \pm 0.0122$, 
and ${\rm PPV} = 0.9444 \pm 0.0062$.  However, the difference between 
these results and those for $W=30$ and $N=2$ are insignificant, with 
$ACC=0.9413$.  In fact ${\rm ACC}$ and ${\rm MCC}$ remain above 0.93 
and 0.86, respectively, over full ranges $N$ and $W$, and Naive Bayes
performance, best at $W=30$, is again statistically indistinguishable 
from that at $N=2$ with $W=10$.

\item {\bf Procedure 6: P6}  As a final exercise in model development,
we pursue the option of working with a binary representation of the
inputs to the HOPP networks constructed.  Each value of the 30 measured
nuclear features was approximiated by a $B$-bit input, with experiments
performed for $B =1,$ 2, and 3.  For each choice of $B$, correlation
orders $N$ up to 4 were studied, with the maximum number of weights
$W$ ranging from 50 to 190 in steps of 10 for each $N$.  The ``best'' 
results for the test-set ensembles achieved in this exercise (otherwise 
conducted as in the procedures above) were found for $B=4$, $N=2$ and 
$W=120$, with ${\rm ACC} = 0.9438\pm 0.0314$, ${\rm MCC} = 0.8786 
\pm 0.0682$, ${\rm SENS} = 0.9618 \pm 0.0320$, ${\rm SPEC} = 0.9129 
\pm 0.0669$, and ${\rm PPV} = 0.9495 \pm 0.0375$.  Marginally different 
results were found (i) at $B = 4$, $N=2$, $W=180$, $B=4$, $W=N$, 
$W=150$, (ii) $B=3$, $N=2$, $W=180$, and (iv) $B=3$, $N=2$, $W=170$.  
The corresponding training-set results for the ``best'' case are 
perfect; the same is true for cases (i)-(iv).  It should be emphasized 
that many other ensembles show performance just as good, within the noise 
measures.  Quite striking results were obtained for the special case 
of an ensemble of Naive Bayes networks in which the inputs are
represented by single bits, with all adjustable weights eligible,
marginally better than all the other cases: ${\rm ACC} = 0.9453\pm 0.0288$, 
${\rm MCC} = 0.8830 \pm 0.0614$, ${\rm SENS} = 0.9673 \pm 0.0318$, 
${\rm SPEC} = 0.9085 \pm 0.0645$, and ${\rm PPV} = 0.9468 \pm 0.0374$.  
Again, training-set performance was perfect.

This last alternative is of special significance from the methodogical 
standpoint.  With binary rather than continuous inputs, the resulting
HOPP models possess both the architecture and input coding required
for rigorous achievement of Bayes inference to given order $N$
the expansion (\ref{uexp}) in irreducible correlatioms among input 
variables.  Aside from truncation of this expansion at finite order, 
the only departure from true optimal Bayesian performance stems from 
the limitations of the gradient-descent learning algorithm adopted
for determination of the set of connection weights.
\end{itemize}

\section{Discussion and Conclusion}

For given input data, the output of the HOPP neural network provides, 
by design, a probabilistic estimate that the tumor is malignant along
with complementary probability that it is benign.  It does not make a 
yes-no decision; rather, it yields further information to help the 
clinician to make a reasoned human decision.  In this respect, the 
strongly bivalent behavior of the probability of malignancy shown in
Fig.~8 is especially revealing.  From this behavior and other
aspects of the problem that have been uncovered in the development
of HOPP models for outcome prediction, the measured features of the 
Wisconsin Breast Cancer Database (WBCD) {\it overdetermine} the correct 
decision. Quite obviously, the redundancy among the set of features 
used as inputs implies that a smaller subset of them and/or their mutual 
correlations may well be sufficient for reliable diagnosis in clinical 
practice, when supplemented by the judgment of the physician.  Granted this 
conclusion, the clear objective of future work should be to identify those 
minimal sets features and their combinations, which maintains maximal 
performance in identification of cancerous biopsies.

\begin{figure} 
\includegraphics[width=.45\textwidth]{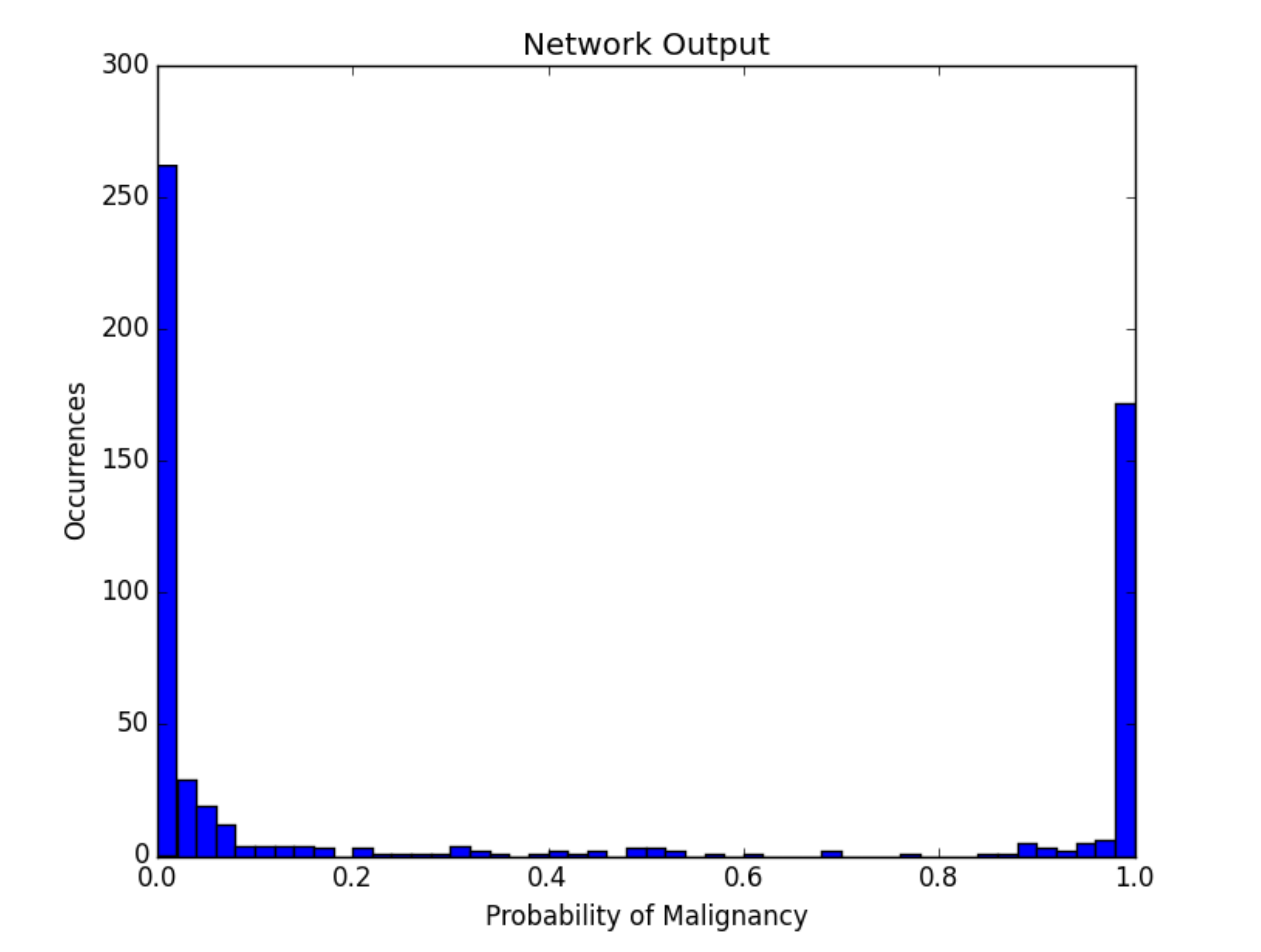}
\begin{center}
\caption{Application of the network described in this paper generates an 
estimate of the probability of the tumor being benign or malignant. The 
frequency of occurrence of the different probability values for the whole 
data set is shown, with the resulting distribution peaking sharply at the 
definite benign and malignant assignments. The thin population in the middle 
ground may correspond to cases in which the disease is not completely 
expressed. The results shown were obtained for a typical network trained on 
all 30 measured features.}
\label{Malignancy-Distribution}
\end{center}
\end{figure}

The results of an initial step toward this goal are 
summarized in Table 1 and its caption, based on the frequency
with which the different factors (biopsy features and their 
combinations) survive the culling process in the creation of
300 candidate HOPP classifiers.

Given a database of measured physical characteristics (``features'') of 
the nuclei of cells extracted through Fine Needle Aspiration of breast 
cancer tumors, a procedure has been developed that not only provides for
identification of those features most essential for correct diagnosis 
of a tumor, but also opens the prospect of finding those {\it combinations} 
of features that are most salient, up to arbitrary order.  This program 
was implemented by extending the Higher-Order Probabilistic Perceptron 
(HOPP) model of Clark et al.~[10] to continuous inputs and including, 
in its two-layer feedforward architecture, higher-order correlations among 
input variables up to fourth order.  The resulting classification 
scheme, including 30 third-order terms, yielded a Matthews coefficient of 
0.90 and an average predictive accuracy of 97\%, with minimal
deviations from this average performance from one model realization
to another. These results are indicative of successful and reliable 
classification, with minimal sampling bias. A remarkable finding
is the surprisingly high quality of performance already attained at 
the Naive Bayes level (first order in the expansion of the log of the 
predicted probability distribution).  This suggests that the HOPP
approach, designed to introduce effects of higher order correlations
among input variables, amounts to ``overkill'' in this iconic 
machine-learning problem.  The distinct advantages of the HOPP
design will become more apparent in application to more challenging
problems of classification and regression.

\section{Acknowledgments}
This research was supported in part by the McDonnell Center for the Space
Sciences.  JWC is grateful for the generous hospitality of the Center for 
Mathematical Sciences of the University of Madeira.

\section{Appendix: Embedding Bayes' Rule in a Neural Network}

An important set of problems in pattern recognition involves the 
classification of input patterns represented by vectors 
$x=(x_1,...,x_K)$ of finite length $K$.  It is the task of a 
Classifier System to correctly assign each such input
pattern to one of a finite number $L$ of distinct categories 
$\lambda$ (or in general to one or more categories).

It is well known \cite{duda} that Bayes' rule provides a basis of 
an optimal strategy for dealing with this decision problem.  One 
statement of Bayes' rule (``Bayes theorem of inverse probabilities''),
which follows from the definition of conditional probabilities:

\begin{itemize}
\item[]
{\it The probability of an Hypothesis given the available Data is equal
to the probability of the Data given the Hypothesis (``the likelihood'') 
times the a priori probability of the Hypothesis, divided 
by the a priori probability of the Data.} 
\end{itemize}

Applying Bayes' rule of inference to the stated classification
problem, we may assert that the posterior probability that the
correct category is $\lambda$ when then the input pattern is known
to be $x$ is given by
\begin{equation}
p(\lambda|x) = {{p(x|\lambda)P(\lambda)} \over {p(x)}},  
\end{equation}
where $p(x| \lambda)$ is the class-conditional probability that the 
pattern is $x$ when the category is known to be $\lambda$ 
(the ``likelihood'' of the data $x$) and $P(\lambda)$ is the 
prior probability of finding $\lambda$.  (The denominator 
$p(x)= \sum_{\nu} p(x|\nu)P(\nu)$ is just a normalization constant 
ensuring that $p(\lambda|x) \geq 0 $ defines a probability 
distribution over exhaustive outcomes $\lambda$.)  

The {\it Bayes-optimal decision strategy} is then to select that 
category $\mu$ among $\lambda=1,...,L$ for which the posterior probability
$p(\lambda |x)$ takes its largest value. This recipe is optimal 
in the sense that it minimizes the probability of misclassification.

In the simplest situation, the input variables $x_1$, $x_2$,...,
$x_K$ are independent of one another.  This implies that the 
class-conditional probability $p(x|\lambda)$ must be just a product 
of independent factors $\rho_i(x_i|\lambda)$, one for each variable 
$x_i$.  However, in principle and in practice, correlations between 
these variables of all orders up to $K$ may be present in the vectors 
of the data ensemble.  This general situation can be represented 
by the product decomposition
\begin{eqnarray}
p(x|\lambda) &=& \prod_{i}\rho_i(x_i|\lambda) 
\prod_{i<j} \rho_{ij}(x_i x_j|\lambda) 
\prod_{i<j<k} \rho_{ijk}(x_ix_jx_k|\lambda) \nonumber \\
&&\cdots \rho_{1...K} (x_1...x_K|\lambda), 
\label{prod}
\end{eqnarray}
where the non-negative factors $\rho_{\ldots}$ together maintain
the requirement $0 \leq p(x|\lambda) \leq 1$.  The $K!/m!(K-m)!$ 
factors $\rho_{k_1 \cdots k_m} (x_{k_1}\ldots x_{k_m}|\lambda)$ 
arising for a given order $m$ may be different functions
of their arguments.  

\begin{itemize}
\item[(i)]
The first product, over the $\rho_i(x_i,\lambda)$, represents
the contribution to the class-conditional probability from
the inputs $x_i$ regarded as independent variables.
\item[(ii)]
The second product, over the $\rho_{ij}(x_i,x_j)$, represents the
modification produced by pairwise correlations of the distinct 
inputs $x_i$ and $x_j$.
\item[(iii)]
The third product represents the modification due to irreducible triplet 
correlations of the distinct inputs $x_i$, $x_j$, $x_k$.
\item[(iv)]
And so on to the effects of irreducible $K$-wise correlations.  
\end{itemize}

Whether the product decomposition (or identity) is useful in practice 
depends on the complexity (or simplicity) of the decision tree 
of the classification problem.  It will be useful if $p(x|\lambda)$ 
can be accurately approximated by the nontrivial inclusion of a 
relatively small number of factors in the general product, with 
the other factors set to their trivial values of unity.  Keeping 
only the first $\prod$ factor, i.e., assuming the input variables 
to be independent of one another, defines the {\it Naive Bayes 
Classifier} \cite{duda,mpap}.  Quite evidently, other representations
of the posterior probability allowing for correlations among 
input variables are available and may prove more or less advantageous
depending on the classification problem to be solved, one example
being the restricted product representation adopted by Duda and Hart
\cite{duda} to generate the Chow expansion.  There are interesting
analogies to the structure of variational wave functions employed
in {\it ab initio} theories of quantum many-particle systems to
include effects of two-particle, three-particle, and multi-particle 
correlations of higher orders.

The simple steps above have exposed the essential structural requirements 
for a Bayes classifier in terms of an exact product decomposition of 
the class-conditional probability. We are now prepared to establish 
a precise connection with neural-network architecture.  Consider a 
two-layer feedforward neural network consisting of:
(i) $K$ input neurons (labeled $i$,$j$,$k$, etc.\ or $i_1,i_2,\ldots,i_K$) 
whose activities register the components of a given pattern vector $x$ 
and (ii) $L$ output neurons $\lambda$ whose activities $y_\lambda$ are to 
be interpreted as probabilities that the input belongs to 
the corresponding classes labeled $\lambda_1,\ldots,\lambda_L$.

To guarantee that the outputs $y_\lambda$ are positive and sum to 
unity, as required for a probability distribution, assume them to 
be given by the ``soft-max'' squashing function 
\begin{equation}
y_\lambda(x) = 
{{{Eq.~\rm e}^{u_\lambda(x)}} \over {\sum_\nu {\rm e}^{u_\nu (x)}}},
\end{equation}
where $u_\lambda$ is the net stimulus to unit $\lambda$
from all the units of the input layer.  The nature of the interactions 
between input and output neurons is still to be specified. 

Thus we come to the crucial step in the constructive proof of the
structural correspondence between Bayes Rule and a two-layer feedforward
neural network.  The Bayes posterior probability for each class 
$\lambda$ is identified with the activity $y_\lambda$ of the output 
neuron corresponding to that class:
\begin{equation}
p(\lambda|x) = {{p(x|\lambda)P(\lambda)} \over {\sum_\nu p(x|\nu)P(\nu)}}
=y_\lambda(x) \equiv
{{{\rm e}^{u_\lambda(x)}} \over {\sum_\nu {\rm e}^{u_\nu(x)}}} 
\end{equation}
Asserting the obvious identification
\begin{equation}
u_{\lambda}(x)= \ln \bigl[ p(x|\lambda)P(\lambda) \bigr] = \ln p(x|\lambda)
+ \ln P(\lambda), 
\end{equation}
the quantity $\ln p(x|\lambda)$ is readily from the product representation,
Eq.~(\ref{prod}):
\begin{eqnarray}
\ln p(x|\lambda) &=& \sum_i \ln \rho_i(x_i|\lambda) 
+ \sum_{i<j} \ln \rho_{ij}(x_ix_j|\lambda) \nonumber \\
&+& \sum_{i<j<k} \ln \rho_{ijk}(x_ix_jx_k|\lambda) \\
&+& \cdots + \ln\rho_{1\cdots K}(x_1...x_K|\lambda) .
\label{sumoflogs}
\end{eqnarray}

The next task to reduce this sum-of-logs expression for the 
posterior probability and translate the $u_\lambda$ identification 
into a more familiar form for the stimulus as a sum of weighted 
neuronal activities.  To do so, we restrict considerations to 
the case that the input patterns are bit strings, i.e.\ $x_k 
\in \{0,1\}$, with $k=1,...,K$.   Writing $1-x_k= {\bar x}_k$ 
we introduce an abbreviated notation through the examples 
\begin{equation}
\rho(x_1=1,x_2=0 | \lambda )=\rho_{1 {\bar 2} , \lambda } 
\end{equation}
and
\begin{equation}
\rho(x_1=1, x_2=1, x_3=0, x_4=1 | \lambda) =\rho_{1 2 {\bar 3} 4, 
\lambda } .
\end{equation}

Next we generate a sequence of identities beginning with
\begin{equation}
\rho(x_1x_2 | \lambda ) = \rho_{12,\lambda}^{x_1x_2}
\rho_{1 {\bar 2} ,\lambda}^{x_1{\bar x}_2}
\rho_{{\bar 1} 2,\lambda}^{{\bar x}_1x_2}
\rho_{{\bar 1} {\bar 2} ,\lambda}^{{\bar x}_1{\bar x}_2} 
\end{equation}
at the level of two correlated inputs, then to
\begin{eqnarray}
\rho(x_1x_2x_3| \lambda ) &=&
\rho_{123,\lambda}^{x_1x_2x_3}
\rho_{12 {\bar 3},\lambda}^{x_1x_2{\bar x}_3}
\rho_{1 {\bar 2} {\bar 3},\lambda}^{x_1{\bar x}_2x_3}
\rho_{{\bar 1} 23,\lambda}^{{\bar x_1}x_2x_3} 
\rho_{1 {\bar 2} {\bar 3},\lambda}^{x_1{\bar x}_2{\bar x}_3} \nonumber \\
&\times&
\rho_{{\bar 1} 2 {\bar 3},\lambda}^{{\bar x}_1x_2{\bar x}_3}
\rho_{{\bar 1} {\bar 2} 3,\lambda}^{{\bar x}_1{\bar x}_2x_3} 
\rho_{{\bar 1} {\bar 2} {\bar 3},\lambda}^{{\bar x}_1{\bar x}_2{\bar x}_3}
\end{eqnarray}
for three, and so on for combinations of four inputs, etc., and
and ending with $\rho(x_1 \cdots x_K|\lambda)$.

Introducing these identities into the sum-of-logs formula
(\ref{sumoflogs}) for the posterior probability, straightforward
manipulations yield an expression for the stimulus of the output 
neuron $\lambda$ having the form
\begin{eqnarray}
u_{\lambda}(x) &=& w_{\lambda,0} + \sum_i w_{\lambda,i} x_i 
+ \sum_{i<j} w_{\lambda,ij} x_i x_j \nonumber \\
&+& \sum_{i<j<k} w_{\lambda,ijk} x_i x_j x_k + \cdots +
\nonumber \\
&+& \cdots + w_{\lambda,12 \cdot \cdot K} x_1 \cdots x_K  .
\end{eqnarray}
The sums are performed over {\it distinct} input indices 1 through
$K$.  Explicit formulas may be given for the connection weights 
$ w_{\lambda, k_1...k_m}$ and biases $w_{\lambda,0}$ in terms of 
the conditionals $\rho(x_{k_1}\ldots x_{k_m}|\lambda)$ and 
the prior $P(\lambda)$, for $\lambda = 1, 2, \ldots L$.  The 
familiar architecture of Rosenblatt's Elementary Perceptron 
\cite{rosenblatt}, involving only pairwise couplings from input 
to output neurons, is recovered upon suppression of all terms 
nonlinear in the input variables.

We conclude that, in general, the two-layer probabilistic network 
with soft-max output functions can match Bayes' rule if and only 
if forward couplings from the input neurons to the output neurons 
of all orders up to $K$ are permitted.  Such a neural network is
called a Higher-Order Probabilistic Perceptron (HOPP).  Importantly,
the above proof that the HOPP architecture is general enough to 
realize Bayes-optimal inference is based on the assumption of binary 
inputs, $x_i \in \{0,1\}$.  This desirable feature need not hold 
otherwise, and in particular for the floating-point inputs involved
in the WBCD machine-learning problem.

Naturally, architecture is only part of the story, as, in effect it
establishes only the class of functions that is being used to
approximate the actual input-output ``decision'' underlying the 
data at hand.  In real-world problems one hardly ever has direct 
access to the conditionals and priors needed to evaluate the weights 
and biases that describe the interactions experienced by the surrogate 
neuronal units.  In the traditional neural network context, they 
must be learned from from examples.  

As one option, it is natural to formulate a procedure for training 
HOPPs based on incremental gradient-descent minimization of
a squared-error cost function 
\begin{equation}
C = {1 \over 2}\sum_\lambda^L (t_\lambda - a_\lambda)^2,
\end{equation}
just as in backpropagation, but without the complication of hidden 
layers.  Alternatively, a relative-entropy cost function 
(Kullback-Leibler entropy) may be used.  It is expected that 
these procedures will move the weight/bias parameters ever 
closer to those needed to match Bayes performance, or to
achieve some other optimization goal, but the sense and 
degree that this is being accomplished are in general elusive.

At this point, theorems established by Ruck et~al. \cite{ruck}, 
Wan \cite{wan}, and Richard and Lippmann \cite{richard} help clarify 
the situation.  Two special cases are considered:
\begin{itemize}
\item[(i)]
The usual case where the target (desired) outputs are ``1 of $L$'', 
meaning that one output should be unity, and all others zero.
\item[(ii)]
In the other case, the target outputs are binary, unity
or zero, without regard for the number that can simultaneously
take the value 1.  
\end{itemize}
Inputs may be continuous or binary.  To paraphrase Richard and 
Lippmann (1991):
\begin{itemize}
\item[(i)]
When network parameters [i.e., the weights and biases] are
chosen to minimize a squared-error cost function, outputs estimate
the conditional expectations of the desired outputs so as to
minimize the mean-squared estimation error.  
\item[(ii)]
For a 1-of-$L$ problem, when network parameters are chosen to 
minimize a squared-error cost function, the outputs provide direct
estimates of the posterior probabilities of the Bayes classifier
so as to minimize the mean-squared estimation error.  
\item[(iii)]
When the desired outputs are binary but not necessarily 
1-of-$L$ and the network parameters are chosen to minimize a squared-error 
cost function, the outputs estimate the conditional probabilities that 
the desired outputs are unity, given the input.
\end{itemize}


\begin{thebibliography}{99}
	
\bibitem{bcstatis}
\url{https://www.breastcancer.org}

\bibitem{yu}
Y.-H. Yu, W. Wei, and J.-L. Liu, Diagnostic value of fine-needle 
aspiration biopsy for breast mass: a systematic review and meta-analysis, 
BMC Cancer, 25 January 2012, 12-41. 

\bibitem{wolberg90}
W. H. Wolberg and O. L. Mangasarian, Multisurface method of pattern 
separation for medical diagnosis applied to breast cancer cytologies 
Proc. Nat. Acad. Sci. USA 87 (1990) 9193-9196. 

\bibitem{wolberg93a}
W. H. Wolberg, W. N. Street, and O. L. Mangasarian, Breast cytology
diagnosis via digital image analysis, Anal. Quant. Cytol. Histol. 15,
(1993) 396-404).

\bibitem{wolberg93b}
W. N. Street, W. H. Wolberg, and O. L. Mangasarian,  Nuclear feature
extraction for breast tumor diagnosis, Proc.\ SPIE 1905, Biomedical Image 
Processing and Biomedical Visualization, 861-870 (1993).

\bibitem{wolberg94}
W. H. Wolberg, W. N. Street, and O. L. Mangasarian, Machine 
learning techniques to diagnose breast cancer from fine-needle 
aspirates, Cancer Letters 77 (1994) 163-171. 

\bibitem{wolberg95a}
O. L. Mangasarian, W. N. Street and W. H. Wolberg, Breast cancer diagnosis
and prognosis via linear programming, Operations Research 43 (1995) 570-577. 

\bibitem{wolberg95b}
W. H. Wolberg, W. N. Street, and O. L. Mangasarian, Image analysis and 
machine learning applied to breast cancer diagnosis and prognosis, 
Anal. Quant. Cytol. Histol. 17 (1985) 77-87.

\bibitem{mang68} 
O. L. Mangasarian, Multi-surface method of pattern separation, IEEE 
Trans.\ on Information Theory, IT-14 (1968) 801-807.

\bibitem{mang93} 
O. L. Mangasarian, Mathematical programming in neural networks, ORSA Journal 
on Computing 5 (1993) 349-360. 

\bibitem{bennett92a}
K. P. Bennett, Decision tree construction via linear programming,
in M. Evans, ed., {\it Proceedings of the 14th Midwest Artificial 
Intelligence and Cognitive Science Society Conference} (1992),
pp.\ 97-101.

\bibitem{bennett92b}
K. P. Bennett and O. L. Mangasarian, Robust linear programming 
discrimination of two linearly inseparable sets, in {\it Optimization 
Methods and Software} 1 (1992), 23-34.

\bibitem{dataset}
archive.ics.uci.edu/ml/datasets/Breast+Cancer+Wisconsin +(Original),
UCI Machine Learning Repository, Irvine, CA: University of California, 
School of Information and Computer Science
[\url{https://archive.ics.uci.edu/ml}]. 

\bibitem{dawson}
A. E. Dawson, R. E. Austin, D. S. Weinberg, Nuclear grading of breast 
carcinoma by image analysis. Am. J. Clin. Pathol. 95 (1991) S29-37).

\bibitem{maclin1}
P. S. Maclin, J. Dempsey, J. Brooks, and J. Rand,  Using neural 
networks to diagnose cancer. J. Med. Syst. 15 (1991) 11-19.

\bibitem{maclin2}
P. S. Maclin and J. Dempsey,  Using an artificial neural network to 
diagnose hepatic masses, J. Med. Syst. 16 (1992) 215-225.

\bibitem{golberg}
V. Golberg, A. Manduca, D. L. Ewert, J. J. Gisvold, J. F. Greenleaf,
Improvement in specificity of ultra-sonography for diagnosis of breast 
tumors by means of artificial intelligence, Med. Phys. 19 (1992) 
1475-1481.

\bibitem{wilding1}
M. L. Astion, P. Wilding, Application of neural networks to the 
interpretation of laboratory data in cancer diagnosis, Clin. Chem. 38 
(1992) 34-38.

\bibitem{wilding2}
P. Wilding, M. A. Morgan, A. E. Grygotis, M. A. Shoffner, E. F. Rosato, 
Application of backpropagation neural networks to diagnosis of breast 
and ovarian cancer, Cancer Letters 77 (1994) 145-153.

\bibitem{ravdin1}
P. M. Ravdin, G. M. Clark, S. G. Hilsenbeck, M. A. Owens. M.A., P. Vendely,
M. R. Pandian, and W. L. McGuire. W.L., A demonstration that breast 
cancer recurrence can be predicted by neural network analysis, 
Breast Cancer Res. Treat. 21 (1992) 47-53. 

\bibitem{ravdin2}
P. M. Ravdin and G. M. Clark, A practical application of neural network 
analysis for predicting outcome of individual breast cancer patients,
Breast Cancer Res. Treat. 22 (1992) 285-293.

\bibitem{ravdin3}
W. L. McGuire, A. K. Tandon, D. C. Allred, G. C. Chamness, P. M. Ravdin, 
and G. M. Clark, Treatment decisions in axillary node-negative breast 
cancer patients, J. Natl. Cancer Inst. Monogr. II (1992) 173-180.

\bibitem{nafe}
R. Nafe and H. Choritz (1992) Introduction of a neuronal network as a 
tool for diagnostic analysis and classification based on experimental 
pathologic data, Exp. Toxicol. Pathol. 44 (1992) 17-24.

\bibitem{wu}
Y. Wu, M. L. Giger, K. Doi, C. J. Vyborny. R. A. Schmidt, and C. E. Metz, 
Artificial neural networks in mammography: Application to decision 
making in the diagnosis of breast cancer, Radiology 187 (1993) 81-87.

\bibitem{rogers}
S. K. Rogers, D. W. Ruck, and M. Kabrisky, Artificial neural networks for 
early detection and diagnosis of cancer, Cancer Letters 77 (1994) 79-83.

\bibitem{floyd}
C. E. Floyd, J. Y. Lo, A. J. Yun, D. C. Sullivan, and P. J. Kornguth,
Prediction of breast cancer malignancy using an artificial neural network,
Cancer 74 (1994) 2944-2948.

\bibitem{roy}
A. Roy, S. Govil, and R. Miranda, An algorithm to generate Radial Basis 
Function (RBF)-like nets for classification problems, Neural Networks 8
(1995) 179-201.

\bibitem{fogel1}
D. B. Fogel, E. C. Wasson, and E. M. Boughton, Evolving neural networks 
for detecting breast cancer, Cancer Letters 96 (1995) 49-53. 

\bibitem{fogel2}
D. B. Fogel, E. C. Wasson EC, and V. W. Porto, A step toward computer-assisted 
mammography using evolutionary programming and neural networks, 
Cancer Letters 119 (1997) 93-97.

\bibitem{sahiner}
B. Sahiner, H.-P. Chan, N. Petrick, D. Wei, M. A. Helvie, D. D. Adler, 
and M. M. Goodsitt, Classification of mass and normal breast tissue: 
A convolution neural network classifier with spatial domain and 
texture images, IEEE Transactions on Medical Imaging 15 (1996) 598-610. 

\bibitem{setiono}
R. Setiono, Extracting rules from pruned neural networks for breast
cancer diagnosis, Artificial Intelligence in Medicine 8 (1996) 37-51.

\bibitem{burke}
H. B. Burke, P. H. Goodman, D. B. Rosen, D. E. Henson, J. N. Weinstein, 
F. E. Harrell, J. J. R. Marks, D. P. Winchester, D. G. Bostwick, 
Artificial Neural Networks improve the accuracy of cancer survival prediction, 
Cancer 79 (1997) 857-862.

\bibitem{furundzic}
D. Furundzic, M. Djordjevic, and A. J. Bekic, Neural networks approach 
to early breast cancer detection, Syst. Architect. 44 (1998) 617-633.

\bibitem{pena}
C. A. Pe\~na-Reyes and M. Sipper, A fuzzy-genetic approach to 
breast cancer diagnosis, Artificial Intelligence in Medicine 17 (1999) 
131-155. 

\bibitem{setiono2}
R. Setiono, Generating concise and accurate classification rules for breast 
cancer diagnosis, Artificial Intelligence in Medicine 18 (2000) 205-219.

\bibitem{west}
D. West and V. West Model selection for a medical diagnostic decision support 
system: a breast cancer detection case, Artificial Intelligence in Medicine 
20 (2000) 183-204.

\bibitem{abbass}
H. A. Abbass, An evolutionary artificial networks approach to breast cancer,
Artificial Intelligence in Medicine 25 (2002) 265-281.

\bibitem{aragones}
M. J. Aragones, A. G. Ruiz, R. Jimenez, M. Perez, and E. A. Conejo,
A combined neural network and decision trees model for prognosis 
of breast cancer relapse, Artificial Intelligence in Medicine 27 
(2003) 45-53.

\bibitem{meesad}
P. Meesad and G. Yen, Combined numerical and linguistic knowledge 
representation and its application to medical diagnosis, IEEE 
Transactions on Systems, Man, and Cybernetics, Part A: Systems and 
Humans 3 (2003) 206-222. 

\bibitem{chen}
C. M. Chen, Y. H. Chou, K. C. Han, G. S. Hung, C. M. Tiu, H. J. Chiou,
and S. Y. Chiou, Breast lesions on sonograms: Computer aided diagnosis 
with nearly setting-independent features and Artificial Neural 
Networks, Radiology 226 (2003) 504-514. 

\bibitem{kiyan}
T. Kiyan and T. Yildirlm, Breast cancer diagnosis using statistical 
neural networks, Journal of Electrical and Electronics Engineering 14 
(2004) 1149-1153. 

\bibitem{chou}
S. M. Chou, T. S. Lee, Y. E. Shao, and I. F. Chen,
Mining the breast cancer pattern using Artificial Neural Networks and 
multivariate adaptive regression splines, Expert Syst. Applications 
27 (2004) 133-142.

\bibitem{tcchen}
A GAs based approach for mining breast cancer pattern, T.-C. Chen and 
T.-C. Hsu, Expert Syst. Applications 30 (2006) 674-681.

\bibitem{nahar}
J. Nahar, Y. P. P. Chen, and S. Ali, Kernel based Naive Bayes Classifier 
for breast cancer prediction, Journal of Biological Systems, 15 (2007) 
17–25.

\bibitem{elizondo}
D. A. Elizondo, R. Birkenhead, M. G\'ongora, E. Taillard, P. Luyima, 
Analysis and test of efficient methods for building recursive deterministic 
perceptron neural networks, Neural Networks 20 (2007) 1095-1108.

\bibitem{uberli}
E. D {\"U}beyli, Implementing automated diagnostic systems for breast cancer 
predictions, Expert Syst. Applications 33 (2007) 1054-1062.

\bibitem{jelen}
L. Jelen, T. Fevens, and A. Krzyzak, Classification of breast cancer 
malignancy using cytological images of Fine Needle Aspiration Biopsies, 
International Journal of Applied Mathematics and Computer Science 18 
(2008) 75-83.

\bibitem{huang}
C. L. Huang, H.-C. Liao, and M.-C. Chen, M.-C. (2008). Prediction, 
model building and feature selection with support vector machines 
in breast cancer diagnosis, Expert Syst. Applications 34 (2008) 578–587. 

\bibitem{akay}
M. F. Akay, SVMs combined with feature selection for breast cancer
diagnosis, Expert Syst. Applications 36 (2009) 3240-3247.

\bibitem{subashini}
T. S. Subashini, V. Ramalingam, and S. Palanivel,
Breast mass classification based on cytological patterns using RBFNN 
and SVM, Expert Syst. Applications 36 (2009) 5284-5290.

\bibitem{karabatak}
M. Karabatak and M. C. Ince, An expert system for detection of 
breast cancer based on association rules and neural network, Expert 
Syst. Applications 36 (2009) 3465.

\bibitem{liang}
L. Liang, F. Cai, and Vladimir Cherkassky, Predictive learning with 
structured (grouped) data, Neural Networks 22 (2009) 766-773.

\bibitem{paulin09}
F. Paulin and A. Santhakumaran, Extracting rules from feed forward neural 
networks for diagnosing breast cancer, CiiT International Journal of 
Artificial Intelligent Systems and Machine Learning 1 (2009) 143-146.

\bibitem{paulin10}
F. Paulin and A. Santhakumaran, Back propagation neural network by 
comparing hidden neurons: Case study on breast cancer diagnosis,
International Journal of Computer Applications (0975 – 8887) 2 (2010) 
40-44.

\bibitem{paulin11}
F. Paulin and A. Santhakumaran, Classification of breast cancer by
comparing back propagation training algorithms, International Journal 
on Computer Science and Engineering 3 (2011) 327-332.

\bibitem{peres}
R. T. Peres and C. E. Pedreira, A new local–global approach for 
classification, Neural Networks 23 (2010) 887-891.

\bibitem{moradi}
M. Fallahnezhad, M. H. Moradi, and S. Zaferanlouei, A Hybrid Higher 
Order Neural Classifier for handling classification problems, 
Expert Syst. Applications 38 (2011) 386-393).

\bibitem{hlchen}
H.-L. Chen, B. Yang, J. Liu, and D.-Y. Liu,
A support vector machine classifier with rough set-based feature 
selection for breast cancer diagnosis, Expert Syst. Applications
38 (2011) 9014-9022. 

\bibitem{cedeno}
A. Marcano-Cede\~no, J. Quintanilla-Dom\'inguez, and D. Andina
WBCD breast cancer database classification applying artificial 
metaplasticity neural network, Expert Syst. Applications 38
(2011) 9573-9579.

\bibitem{elgader}
H. A. Abd Elgader and M. H. Hamza, Breast cancer diagnosis using artificial 
intelligence neural networks, J. Sc. Tech. 121 (2011) 159-171. 

\bibitem{kowal}
M. Kowal, P. Filipczuk, A. Obuchowicz, and J. Korbicz, Computer-aided 
diagnosis of breast cancer using Gaussian mixture cytological image 
segmentation, Journal of Medical Informatics \& Technologies 17 (2011) 
257-262. 

\bibitem{blachnik}
M. Blachnik and W. Duch, LVQ algorithm with instance weighting for generation 
of prototype-based rules, Neural Networks 24 (2011) 824-830.  

\bibitem{george} 
Y. M. George, B. M. Elbagoury, H. H. Zayed, and M. I. Roushdy, Breast 
fine needle tumor classification using neural networks, IJCSI International 
Journal of Computer Science Issues 9 (2012) 247-256. 

\bibitem{aloraini}
A. Aloraini, Different machine learning algorithms for breast cancer
diagnosis, International Journal on Artificial Intelligence Applications
3 (2012) 11-29. 

\bibitem{gu}
B. Gu, J.-D. Wang, Y.-C. Yud, G.-S. Zheng, Y.-F. Huang, and T. Xu,
Accurate on-line $\nu$-support vector learning, Neural Networks 27
(2012) 51-59. 

\bibitem{savitha}
R. Savitha, S. Suresh, and N. Sundararajan, 
A meta-cognitive learning algorithm for a Fully Complex-valued Relaxation 
Network, Neural Networks 32 (2012) 209-218.

\bibitem{alkim}
E. Alkim, E. G\"urb\"uz, and E. Kilic,
A fast and adaptive automated disease diagnosis method with an 
innovative neural network model, Neural Networks 33 (2012) 88-96. 

\bibitem{ahchen}
A. H. Chen, and C. Yang, The improvement of breast cancer 
prognosis accuracy from integrated gene expression and clinical data,
Expert Syst. Applications 39 (2012) 4785-4795. 

\bibitem{nahar2}
J. Nahar, I. Tasadduq, K. S. Tickle, A. B. M. Shawkat Ali, Y.-P. P. Chen,
Computational intelligence for microarray data and biomedical image 
analysis for the early diagnosis of breast cancer, Expert Syst. Applic. 39 
(2012) 12371-12377. 

\bibitem{utomo}
C. P. Utomo, A. Kardiana, and R. Yuliwulandari, Breast Cancer diagnosis 
using Artificial Neural Networks with Extreme Learning Techniques, 
International Journal of Advanced Research in Artificial Intelligence 3 
(2014) 10-14.

\bibitem{zheng}
B. Zheng, S. W. Yoon, and S. S. Lam, Breast cancer diagnosis based on 
feature extraction using a hybrid of K-means and support vector machine 
algorithms, Expert Sys. Applications 41 (2014) 1476-1482.

\bibitem{dheeba}
J. Dheeba, N. A. Singh, and S. T. Selvi, Computer-aided detection of 
breast cancer on mammograms: A swarm intelligence optimized wavelet 
neural network approach, Journal of Biomedical Informatics 49
(2014) 45–52.

\bibitem{seera}  
M. Seera and C. P. Lim, A hybrid intelligent system for medical data 
classification, Expert. Sys. Applications 41 (2014) 2239-2249.

\bibitem{bhardwaj}
A. Bhardwaj and A. Tiwari, Breast cancer diagnosis using genetically 
optimized neural network model, Expert Syst. Applications 42 (2015) 
4611-4620.

\bibitem{mert}
A. Mert, N. Z. Kilic, E. Bilgli, and A. Akan, Breast cancer detection
with reduced feature set, Computational and Mathematical Methods
in Medicine (2015) 1-11.

\bibitem{nahato}
K. B. Nahato, H. K. Nehemiah, and A. Kannan, Knowledge mining from clinical 
datasets using rough sets and backpropagation neural network, Computational 
Mathematical Methods in Medicine (2015) 1–13.

\bibitem{karabatak2} 
M. Karabatak, A new classifier for breast cancer detection based on 
Naive Bayesian, Measurement 72 (2015) 32-36.

\bibitem{kim}
S. Kim, Z. Yu, R. M. Kil, and M. Lee, Deep learning of support vector 
machines with class probability output networks, Neural Networks 64
(2015) 19-28 (2015).

\bibitem{abdel}
A. M. Abdel-Zaher and A. M. Eldeib, Breast cancer classification using 
deep belief networks, Expert Syst. Applications 46 (2016) 139-144. 

\bibitem{mohammed}
E. A. Mohammed, C. T. Naugler, and B. H. Far, Breast tumor classification 
using a new OWA operator, Expert Syst. Applications 61 (2016) 302-313.

\bibitem{chaurasia}
V. Chaurasia, Saurabh Pal, and B. B. Tiwari, Prediction of benign and 
malignant breast cancer using data mining techniques, Journal of Algorithms
\& Computational Technology 12 (2018) 119-126.

\bibitem{chao}
P. Chao, T. Mazeri, B. Sun, N. B. Weingartner, and Nussinov, The stochastic 
replica approach to machine learning: Stability and parameter optimization 
[arXiv:1708.05715].

\bibitem{rangayyan}
R. M. Rangayyan, F. J. Aires, and J. E. L. Desautels, A review of 
computer-aided diagnosis of breast cancer: Toward the detection 
of subtle signs, J. Franklin Inst. 344 (2007) 312-348.

\bibitem{sadja}
P. Sadja, Machine Learning for Detection and Diagnosis of Diseases,
Annual Review of Biomedical Engineering 8 (2006) 537-565. 

\bibitem{cruz}
J. A. Cruz and D. S. Wishart,  Applications of machine learning in cancer 
prediction and prognosis, Cancer Informatics 2 (2006) 59-77. 

\bibitem{you}
H. You and G. Rumba, Comparative study of classification techniques on 
breast cancer FNA biopsy data, International Journal of Artificial 
Intelligence and Interactive Multimedia 1 (2010) 6-13. 

\bibitem{beg}
M. M. Beg and M. Jain,
An analysis of the methods employed for breast cancer diagnosis,
International Journal of Research in Computer Science 2 (2012) 25-29.

\bibitem{asri}
H. Asri, H. Mousannif, H. Al Moatassime, and T. Noel, Use of machine
learning algorithms for breast cancer risk prediction and diagnosis,
Procedia Computer Science 83 (2016) 1064-1069.

\bibitem{bazazeh}
D. Bazazeh and R. Shubair, Comparative study of machine learning algorithms for breast cancer detection and diagnosis, in {\it Fifth International Conference
on Electronic Devices, Systems and Applications}, Ras Al Khaimah, United Arab 
Emirates (IEEE, New York, 2016), DOI: 10.1109/ICEDSA.2016.7818560. 

\bibitem{mandal}
S. K. Mandal, Performance analysis of data mining algorithms for breast 
cancer cell detection using Naive Bayes, logistic regression, and 
decision tree, International Journal of Engineering and Computer 
Science 6 (2017) 20388-20391.

\bibitem{yue}
W. Yue, Z. Wang, H. Chen, A. Payne, and X. Liu, Machine learning with
applications in breast cancer diagnosis and prognosis, Designs 2, 13,
(2018) 1-17.

\bibitem{kononrev}
I. Kononenko, Machine learning for medical diagnosis: History, state of
the art \& perspective, Artificial Intelligence in Medicine 23 (2001) 
89-109. 

\bibitem{schwarzer}
W. Schwarzer, W. Vach, and M. Schumacher, On the misuses of artificial 
neural networks for prognostic and diagnostic classification in oncology, 
Statistics in Medicine 19 (2005) 541-561. 

\bibitem{liao}
S.-H. Liao, Expert system methodologies and applications -- a decade
review from 1995 to 2004, Expert Syst. Applications 28 (2005) 93-103.

\bibitem{lisboa}
P. J. Lisboa and A. F.G. Taktak, The use of artificial neural networks 
in decision support in cancer: A systematic review, Neural Networks 19
(2006) 408-415.

\bibitem{duda}
R. O. Duda and P. E. Hart, {\it Pattern Classification and Scene
Analysis} (Wiley \& Sons, New York, 1973). 

\bibitem{bishop}
C. M. Bishop, {\it Neural Networks for Pattern Recognition}
(Clarendon Press, Oxford, 1996).

\bibitem{neal}
R. Neal, {\it Bayesian Learning of Neural Networks} (Springer, New York,
1996).

\bibitem{vapnik}
V. Vapnik, {\it Statistical Learning Theory} (Wiley, New York, 1998).

\bibitem{haykin}
S. Haykin, \textit{Neural Networks: A Comprehensive Foundation} 
(Prentice Hall, Upper Saddle River, NJ, 1999).

\bibitem{elomaa}
T. Elomaa, H. Mannila, and H. Toivonen (Eds.), {\it Machine Learning 
2002}, 13th European Conference on Machine Learning, Helsinki, Finland,
August 2002 Proceedings, Springer Lecture Notes in Artificial
Intelligence (Springer, Berlin, 2002). 

\bibitem{clark}
J. W. Clark, K. A. Gernoth, S. Dittmar, and M. L. Ristig,
Higher-Order probabilistic perceptrons as Bayesian inference engine,
Phys. Rev. E 59 (1999) 661-674.

\bibitem{rosenblatt}
F. Rosenblatt, {\it Principles of Neurodynamics} (Spartan Books,
Washington D.~C., 1962).

\bibitem{clarkbook}
J. W. Clark, M. L. Ristig, and Lindenau, {\it Scientific Applications
of Neural Nets}, Springer Lecture Notes in Physics, Vol. 522
(Springer-Verlag, Berlin, 1999).

\bibitem{mandelbrot}
B. B. Mandelbrot, The Fractal Geometry of Nature (W. H. Freeman, 
New York, 1997).

\bibitem{mpap}
M. L. Minsky and S. A. Papert, {\it Perceptrons}, Expanded Edition 
(MIT Press, Cambridge, MA, 1987).

\bibitem{ruck}
D. W. Ruck, S. K. Rogers, M. Kabrisky, M. E. Oxley, and B. W. Suter, 
The multilayer perceptron as an approximation to a Bayes optimal 
discriminant function, IEEE Trans. Neural Netw. 1 (1990) 296-298.

\bibitem{wan}
E. A. Wan, Neural network classification: a Bayesian interpretation,
IEEE Trans. Neural Netw. 1 (1990) 303-305.

\bibitem{richard}
M. D. Richard and R. P. Lippmann, 
Neural network classifiers estimate Bayesian {\it a posteriori} 
probabilities, Neural Comput. 3 (1991) 461-483. 

\bibitem{kullback}
S. Kullback, {\it Information Theory and Statistics} (Wiley, New York,
NY, 1959).
\end{thebibliography}
\end{document}